\documentclass[acmsmall,authorversion]{acmart}

\setcopyright{acmlicensed}
\acmJournal{PACMHCI}
\acmYear{2022} \acmVolume{6} \acmNumber{ETRA} \acmArticle{146} \acmMonth{5} \acmPrice{15.00}\acmDOI{10.1145/3530887}

\AtBeginDocument{%
  \providecommand\BibTeX{{%
    \normalfont B\kern-0.5em{\scshape i\kern-0.25em b}\kern-0.8em\TeX}}}

\usepackage{subcaption}
\usepackage{cleveref}
\usepackage{hyperref}
\usepackage{tabularx}
\makeatletter
\def\hlinewd#1{%
\noalign{\ifnum0=`}\fi\hrule \@height #1 %
\futurelet\reserved@a\@xhline}
\makeatother

\Crefname{section}{Section}{Section} % Use small cref to print Sec., Fig. Tab.
\crefname{section}{Sec.}{Sec.} % Use small cref to print Sec., Fig. Tab.
\crefname{figure}{Fig.}{Fig.} 
\Crefname{figure}{Figure}{Figure} 
\Crefname{table}{Table}{Table} 
\crefname{table}{Tab.}{Tab.} 
\Crefname{equation}{Equation}{Equation} 
\crefname{equation}{Eqn.}{Eqn.} 

\begin{document}
\title{Where and What: Driver Attention-based Object Detection}

%%
%% The ``author`` command and its associated commands are used to define
%% ``authornote`` and ``authornotemark`` commands
\author{Yao Rong}
\email{yao.rong@uni-tuebingen.de}
\affiliation{%
  \institution{University of T\"ubingen}
  \streetaddress{Sand 14}
  \city{T\"ubingen}
  \country{Germany}
  \postcode{72076}
}

\author{Naemi-Rebecca Kassautzki}
\affiliation{%
  \institution{University of T\"ubingen}
  \streetaddress{Sand 14}
  \city{T\"ubingen}
  \country{Germany}
  \postcode{72076}
  }
\email{naemi-rebecca.kassautzki@student.uni-tuebingen.de}

\author{Wolfgang Fuhl}
\affiliation{%
  \institution{University of T\"ubingen}
  \streetaddress{Sand 14}
  \city{T\"ubingen}
  \country{Germany}
  \postcode{72076}
 }
\email{wolfgang.fuhl@uni-tuebingen.de}

\author{Enkelejda Kasneci}
\affiliation{%
  \institution{University of T\"ubingen}
  \streetaddress{Sand 14}
  \city{T\"ubingen}
  \country{Germany}
  \postcode{72076}
  }
\email{enkelejda.kasneci@uni-tuebingen.de}

\newcommand{\yao}[1]{\textcolor{black}{#1}}
\newcommand{\naemi}[1]{\textcolor{black}{#1}}
\newcommand{\enkelejda}[1]{\textcolor{black}{#1}}
\renewcommand{\shortauthors}{Yao Rong et al.}
\begin{abstract}
% Human drivers use their eye gaze to focus on critical objects and make decisions upon them in driving. 
% To simulate human drivers’ attention and thus benefit the autonomous driving technology,
Human drivers use their attentional mechanisms to focus on critical objects and make decisions while driving. As human attention can be revealed from gaze data, capturing and analyzing gaze information has emerged in recent years to benefit autonomous driving technology. Previous works in this context have primarily aimed at predicting ``where'' human drivers look at and lack knowledge of ``what'' objects drivers focus on. Our work bridges the gap between pixel-level and object-level attention prediction. Specifically, we propose to integrate an attention prediction module into a pretrained object detection framework and predict the attention in a grid-based style. Furthermore, critical objects are recognized based on predicted attended-to areas. We evaluate our proposed method on two driver attention datasets, BDD-A and DR(eye)VE. Our framework achieves competitive state-of-the-art performance in the attention prediction on both pixel-level and object-level but is far more efficient (75.3 GFLOPs less) in computation.

\end{abstract}




% \ccsdesc[500]{Human-centered computing}



\maketitle

\begin{figure}[h]
  \centering
  \includegraphics[width=\linewidth]{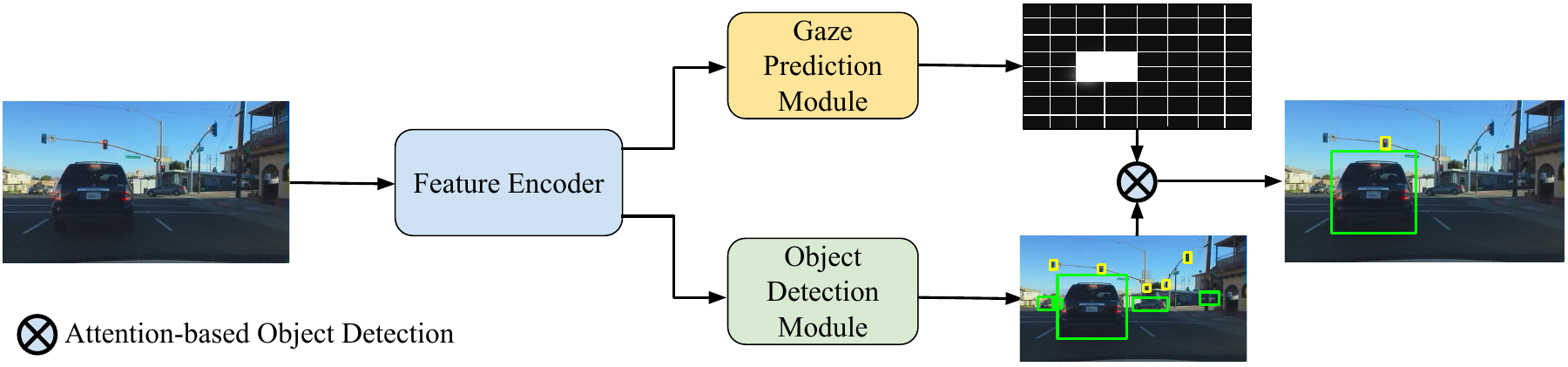}
  \caption{Overview of our proposed critical object detection framework. The \textbf{feature encoder} extracts features from the input image. The \textbf{gaze prediction module} predicts driver attention in a grid-based saliency map and the \textbf{object detection module} detects all the objects in the traffic using extracted features. The \textbf{attention-based objects} are detected and returned to users based on the predicted saliency map and detected objects.}
  \label{fig:teaser}
\end{figure}
\section{Introduction}
%Rewrite this section as: human gaze-based attention in general visual tasks and gaze attention-to-object mapping is important.
\yao{Human attentional mechanisms} play an important role in selecting task-relevant objects effectively in a top-down manner, which can solve the task efficiently \cite{rimey1994control, sutton1988learning, peters2007beyond}. To visualize human attention for these tasks in a general way, a Gaussian filter is applied on fixation points to form a \textit{saliency} map \cite{kummerer2016deepgaze}, thus highlighting the visual attention area. Due to the effectiveness and irreplaceability of human attention in solving visual tasks, visual attention is also being studied in artificial intelligence research (e.g., \cite{zhang2020human}). Many computer vision applications embrace human gaze information, for instance in classification tasks \cite{rong2021human,liu2021goal}, computer-aided medical diagnosis systems \cite{karargyris2021creation,saab2021observational}, or important objects selection/cropping in images and videos \cite{shanmuga2015eye,vasudevan2018object,wang2018salient,santella2006gaze}. %General human attention in driving. 
%For safe driving, efficient visual exploration of the environment is a sensory prerequisite. The visual attention of the driver has been studied extensively in recent decades with the aim of detecting potential driver drowsiness \cite{baccour2019camera, wang2016driver, de2019detection, mandal2016towards, azim2014fully}, recognizing the direction of visual attention \cite{smith2003determining, fridman2015driver, borghi2017poseidon, vicente2015driver,kasneci2017aggregating,kasneci2015online,tafaj2013online,braunagel2017ready}, or anticipating maneuver activities \cite{braunagel2017online, jain2016brain4cars, gebert2019end, rong2020driver, pugeault2015much, kumar2013learning}. 
%Currently, driver attention detection is also used practically in vehicle driver monitoring systems (e.g., \cite{drivermonitoring}) for warning drivers, if they do not concentrate on the traffic.
\yao{To better understand how the human brain processes visual stimuli, knowing not only \textit{where} humans are looking at, but also \textit{what} object is essential, i.e., gaze-object mapping \cite{barz2021visual}. This mapping is needed in many research projects, especially in analytics of student learning process \cite{kumari2021mobile} or human cognitive functions \cite{panetta2019software}.}

% New motivations/usecases
\yao{In autonomous driving applications, successful models should be able to mimic ``gaze-object mapping" of humans, which includes two challenges: Driver gaze prediction and linking the gaze to objects. It is practical to predict driver gaze since sometimes no eye tracker is available or no human driver is required in the higher level of autonomous vehicles. For instance, Pomarjanschi et al. \cite{pomarjanschi2012gaze} validates that highlighting potentially critical objects such as a pedestrian on a head-up display helps to reduce the number of collisions. In this case, a model capable of predicting these critical objects can be used as a ``second driver" and give warnings that assist the real driver. For fully autonomous cars, it is essential to identify these task-relevant objects efficiently to make further decisions and also explain them \cite{kim2018textual}. Recently, there is a growing research interest in predicting human drivers' gaze-based attention \cite{dreyeve2018,xia2018predicting,potentilattention2019}. These existing works predict pixel-level saliency maps, however, they lack semantic meaning of the predicted attention, i.e., the model only predicts \textit{where} drivers pay attention, without knowing \textit{what} objects are inside those areas.} 
\yao{To bridge the research gap between driver gaze prediction and semantic object detection existing in the current research landscape of autonomous driving applications, we propose (1) to predict where and what the drivers look at. Furthermore, we aim (2) at a model that is efficient in computation, since resources on self-driving cars are limited.} Specifically, we designed a novel framework for efficient attention-based object detection based on human driver gaze. Our approach provides not only pixel-level attention saliency maps, but also the information of objects appearing in attention areas, as illustrated in \cref{fig:teaser}. A feature encoder is first used in our framework to encode the information in the input image. Then, the extracted features are used to predict gaze and detect objects in the image at the same time. Since obtaining accurate high-level (object) information is our final goal, instead of low-level (pixel) accuracy in saliency map prediction, we predict salient areas in a grid-based style to save computational costs while still maintaining high performance in the critical object detection task.

%  Our framework combines an object detection network with a light-weight gaze map prediction network. As object detection network we use the state-of-the art architecture YOLO v5 \cite{glenn_jocher_2021_4679653} and compare it with Gaussian YOLO v3 \cite{redmon2018yolov3} and CenterTrack \cite{zhou2020tracking}. While the object detection network is running, we extract feature maps and use them as input for our gaze prediction. 
% Because we do not need perfect prediction on pixel-level, we use a grid as ground-truth, to save computation cost while still maintaining object-level accuracy. For the gaze map prediction, we just use one convolutional layer and one pooling layer to shrink the size and add one dense layer. Therefore, the overall cost is nearly the same as for the object detection network without gaze prediction.

Our contributions can be summarized as follows: (1) We propose a framework to predict objects that human drivers pay attention to while driving. (2) Our proposed grid-based attention prediction module is very flexible and can be incorporated with different object detection models. (3) We evaluate our model on two datasets, BDD-A and DR(eye)VE, showing that our model is computationally more efficient and achieves comparable performance in pixel- and object-level prediction compared to other state-of-the-art driver attention models. For the sake of reproducibility, our code is available at \url{https://github.com/yaorong0921/driver-gaze-yolov5}. 

\section{Related Work}
In the following, \yao{we first discuss previous works of gaze-object mapping used in applications other than driving scenarios and we discuss the novelty of our proposed method for solving this task.} Then, we introduce the related work with a special focus on the driver attention prediction in the context of saliency prediction for human attention, followed by the introduction of several object detectors our framework is based on. \yao{Thanks to deep learning techniques}, there exists a plethora of works in the past decades for visual saliency models and object detectors (see \cite{borji2018saliency,zhao2019object} for review). It is impracticable to thoroughly discuss these works in the two branches, therefore we only present the works which are closely related to our work. %\yao{In the end, we summarize frameworks that aim to automatically map human gaze to attended objects or Area of Interests (AOIs) in different use-cases.}

%Previous works \cite{wolf2018automating,kumar2013learning} intend to avoid massive manual labour for annotating fixation with the object humans are looking at by running automatic gaze-object mapping.
\paragraph{Gaze-Object Mapping.} \yao{Previous works \cite{wolf2018automating,kumar2013learning} set out to reduce tedious labelling by using gaze-object mapping, which annotates objects at the fixation level, i.e., the object being looked at.
One popular algorithm checks whether a fixation lies in the object bounding box predicted by deep neural network-based object detector \cite{kumari2021mobile, barz2021visual,machado2019visual} such as YOLOv4 \cite{bochkovskiy2020yolov4}. Wolf et al.~\cite{wolf2018automating} suggest to use object segmentation using Mask-RCNN \cite{he2017mask} as object area detection. These works train their object detectors with limited object data and classes to be annotated. Panetta et al.~\cite{panetta2019software}, however, choose to utilize a bag-of-visual-words classification model \cite{csurka2004visual} over deep neural networks for object detection due to insufficient training data. 
Barz et al.~\cite{barz2021ar} propose a “cropping-classification” procedure, where a small area centered at the fixation is cropped and then classified by a network pretrained on ImageNet \cite{deng2009imagenet}. This algorithm from \cite{barz2021ar} can be used in Augmented Reality settings for cognition-aware mobile user interaction. In the follow-up work \cite{barz2021visual}, the authors compare the mapping algorithms based on image cropping (IC) with object detectors (OD) in metrics such as precision and recall, and the results show that IC achieves higher precision but lower recall scores compared to OD.}

\yao{
However, these previous works are often limited in object classes and cannot be used to detect objects in autonomous driving applications, since a remote eye tracker providing precious fixation estimation is required for detecting attended objects. Unlike previous gaze-object mapping methods, a model in semi-autonomous driving applications should be able to predict fixation by itself, for instance, giving safety hints at critical traffic objects as a ``second driver" in case human drivers oversee them. In fully autonomous driving, where no human driver fixation is available, a model should mimic human drivers' fixation. Therefore, our framework aims to showcase a driver attention model achieving predicting gaze and mapping gaze to objects simultaneously, which is more practical in autonomous driving applications.}

\paragraph{Gaze-based Driver Attention Prediction.} 
\yao{With the fast-growing interest in (semi-)autonomous driving, studying and predicting human drivers' attention is of growing interest. There are now studies showing improvement in simulated driving scenarios by training models in an end-to-end manner using driver gaze, so that models can observe the traffic as human drivers \cite{liu2019gaze,makrigiorgos2019human}.} Based on new created real-world datasets, such as DR(eye)VE \cite{dreyeve2018} and BDD-A \cite{xia2018predicting}, a variety of deep neural networks are proposed to predict pixel-wise gaze maps of drivers (e.g., \cite{dreyeve2018, xia2018predicting, lv20improving, shirpour2021driver,pal2020looking}). The DR(eye)VE model \cite{dreyeve2018} uses a multi-branch deep architecture with three different pathways for color, motion and semantics. The BDD-A model \cite{xia2018predicting} deploys the features extracted from AlexNet \cite{krizhevsky2012imagenet} and inputs them to several convolutional layers followed by a convolutional LSTM model to predict the gaze maps. An attention model is utilized to predict driver saliency maps for making braking decisions in the context of end-to-end driving in \cite{aksoy2020see}.
Two other well-performing networks for general saliency prediction are ML-Net \cite{cornia2016deep} and PiCANet \cite{liu2018picanet}. ML-Net extracts features from different levels of a CNN and combines the information obtained in the saliency prediction. PiCANet is a pixel-wise contextual attention network that learns to select informative context locations for each pixel to produce more accurate saliency maps. In this work, we will also include these two models trained on driver gaze data in comparison to our proposed model.
%In the context of driver attention, they can only predict which regions are in general often focused by gaze without any driving relevant top-down information. We will use these two networks to compare our results.  
%They use the predicted gaze map for making braking decisions in the context of end-to-end driving. Also, Tawari et al.~\cite{7995828} propose to predict gaze maps with a convolutional neural network, but they use a Bayesian network model. 
Besides these networks, which are focused on predicting the driver gaze map, other models are extended to predict additional driving-relevant areas. While Deng et al.~\cite{potentilattention2019} use a convolutional-deconvolutional neural network (CDNN) and train it on eye tracker data of multiple test person, Pal et al.~\cite{pal2020looking} propose to include distance-based and pedestrian intent-guided semantic information in the ground-truth gaze maps and train models using this ground-truth to enhance the models with semantic knowledge.

%However, all those models were designed for pixel-level prediction of driving relevant areas, thus the output does not contain any information about the relevant objects. 
Nevertheless, these models cannot provide the information of objects that are inside drivers' attention.
It is possible to use the existing networks for detecting attended-to objects, but this would have the disadvantage that predicting gaze maps on pixel-level introduces unnecessary computational overhead if we are just interested in the objects. Hence, going beyond the state of the art, we propose a framework combining gaze prediction and object detection into one network to predict visual saliency in the grid style. Based on a careful experimental evaluation, we illustrate the advantages of our model in having high performance (saliency prediction and object detection) and saving computational resources.

\paragraph{Object Detection.}
In our framework, \yao{we use existing object detection models} for detecting objects in driving scenes and providing feature maps for our gaze prediction module. In the context of object detection, the \textit{You only look once} (YOLO) architecture has played a dominant role in object detection since its first version \cite{redmon2016you}. Due to its speed, robustness and high accuracy, it is also applied frequently in autonomous driving \cite{nugraha2017towards,simony2018complex}. YOLOv5 \cite{glenn_jocher_2021_4679653} is one of the newest YOLO networks that performs very well. Since YOLOv5 differs from traditional YOLO networks and it does not use Darknet anymore, we also consider Gaussian YOLOv3 \cite{choi2019gaussian}. Gaussian YOLOv3 is a variant of YOLOv3 that uses Gaussian parameters for modeling bounding boxes and showed good results on driving datasets. For comparison, we also tried an anchor free object detection network CenterTrack \cite{zhou2020tracking}, which regards objects as points. \yao{By using the feature maps of the object detection network such as YOLOv5 to predict gaze regions, we save the resources of an additional feature extraction module.}

\section{Methodology}
%State-of-the-art driver gaze prediction models are all CNN-based \cite{palazzi2017learning,palazzi2018predicting,xia2018predicting,xia2020periphery,pal2020looking}. The models first extract/encode spatial information into feature maps and then generate/decode these feature maps into gaze maps. In the previous work, the decoder was focusing on predicting attention on a pixel level, requiring a significant amount of computation. Yet, very accurate pixel prediction seems redundant in practical application as long as the attended object can be detected. For this reason, we propose to predict critical regions in a grid. The objects covered by these regions are recognized as 
%\textit{critical objects}. We consider the grid detection problem to be a multi-label classification problem. In this way, we significantly reduce the computation cost without losing performance in critical object detection.
\begin{figure}[h]
  \centering
  \includegraphics[width=\linewidth]{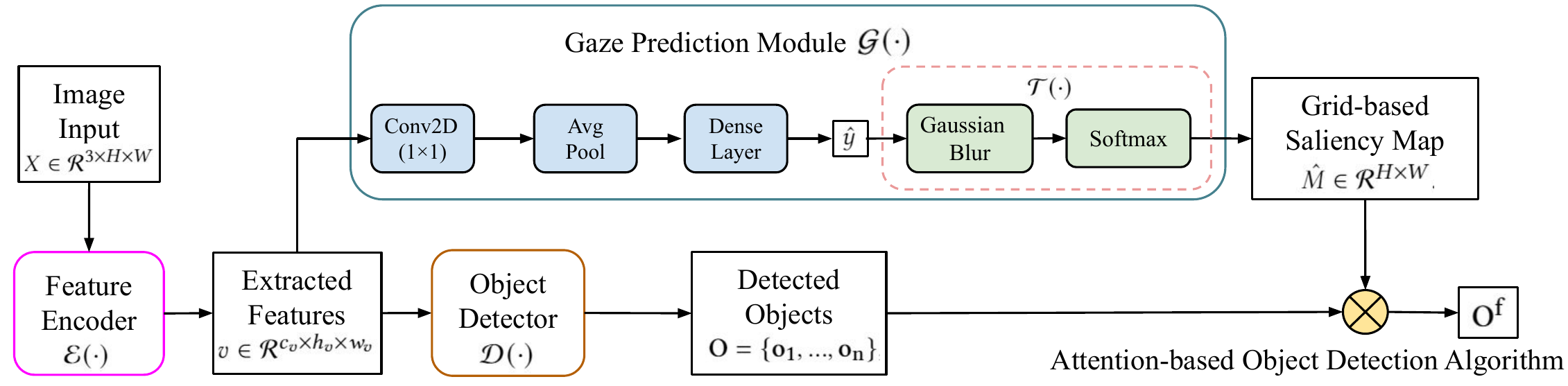}
  \caption{Overview of our proposed driver attention-based object detection framework.}
  \label{fig:detailed framework}
\end{figure}

State-of-the-art driver gaze prediction models extract features from deep neural networks used in image classification or object recognition, e.g., AlexNet \cite{krizhevsky2012imagenet} or VGG \cite{simonyan2014very}, and use decoding modules to predict precise  pixel-level saliency maps. We propose a new approach as shown in \cref{fig:detailed framework} to predict what objects drivers attend to based on a grid-based saliency map prediction. The object detector and attention predictor share the same image features and run simultaneously in a resource-efficient manner. In this section, we first introduce our attention-based object detection framework in \cref{sec:framework}, including the gaze prediction module and object detection algorithm, etc. Implementation details of our model, such as the specific network architecture of network layers are discussed in \cref{sec:implementation}.
% and using an object detection network as image feature extractor for the gaze prediction and simultaneously as detector for the objects. In this section, we first introduce our method for converting gaze maps into grid-vectors as ground-truth and then share our proposed framework.

\subsection{Attention-based Object Detection}
The framework is formalized as follows: Given an RGB image input from driving scenarios $X \in \mathcal{R}^{3 \times {H}\times{W}}$ where $H$ and $W$ refer to the height and width, an image feature encoder $\mathcal{E}(\cdot)$ encodes the input image $X$ into feature $v$. This feature can be a feature map $v \in \mathcal{R}^{c_v \times {h_v}\times{w_v}}$ where $h_v$,$w_v$ and $c_v$ represent the height, width and number of channels of the feature map. $v$ is the input of the gaze prediction module $\mathcal{G}(\cdot)$, which first predicts a grid-vector $\hat{y} = \mathcal{G}(v)$. Then, a transformation operation $\mathcal{T}(\cdot)$ is applied on $\hat{y}$ to turn it into a 2-dimensional saliency map $\hat{M} \in \mathcal{R}^{{H}\times{W}}$. Similarly, the object detection module $\mathcal{D}(\cdot)$ predicts a set of objects appearing in the image $\mathbf{O} = \{\mathbf{o_1}, \mathbf{o_2}, ..., \mathbf{o_n}\}$, where each $\mathbf{o_i}$ contains the bounding box/class information for that object and $n$ is the total number of objects. Based on $\hat{M}$ and $\mathbf{O}$, we run our attention-based object detection operation $\bigotimes$ to get the set of focused objects $\mathbf{O^f}$, which can be denoted as $\hat{M} \bigotimes \mathbf{O} = \mathbf{O^f}$ and $|\mathbf{O^f}|  \leq n$. \cref{fig:detailed framework} demonstrates different modules in our framework.

\begin{figure}[t]
    \centering
    \includegraphics[width=0.7\linewidth]{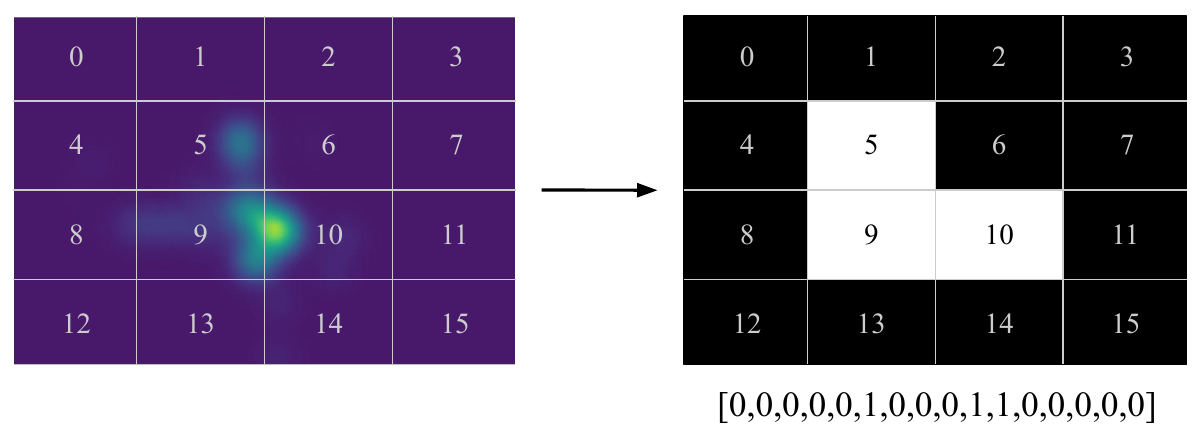}
    \caption{Illustration of transforming a saliency map into a grid-vector. The used grid here is $4\times4$. Grid cells 5, 9 and 10 reach the threshold, therefore the grid-vector $y$ for the saliency map $M$ is $[0,0,0,0,0,1,0,0,0,1,1,0,0,0,0,0]$.}
    \label{fig:grid vector}
\end{figure}

%\subsection{Grid-Vector as Ground-Truth}
\paragraph{Gaze Prediction Module.} To reduce the computational cost, we propose to predict the gaze saliency map in grids, i.e., we alter the saliency map generation problem into a multi-label prediction problem. Concretely, we transform the target saliency map $M \in \mathcal{R}^{{H}\times{W}}$ into a grid-vector $y \in \mathcal{R}^{n\cdot m}$, where $n$ and $m$ are the numbers of grid cells in height and width dimension, respectively.
% Our ground-truth is a vector of dimension $n\times m$ given the grid number $[n,m]$. 
Each entry of the grid-vector $y$ is a binary value. The index of entry corresponds to the index of a region in the gaze map. $1$ means that the region is focused by the driver, while $0$ means not. 
%In driver attention datasets, the ground-truth is a gaze map collected by eye-trackers. 
%(2) Divide the binarized $M$ into regions according to the grid setting ($n$ cells in the width; $m$ cells in the height).
Here, we obtain a grid-vector $y$ from a saliency map $M$ using the following procedure: (1) We binarize the $M$ to $M'$ with a value of 15\% of the maximal pixel value (values larger than it will be set to 1, otherwise to 0). (2) For each grid cell ($j$-th entry in the $y$), we assign a ``probability'' of being focused as $p = \frac{\sum{M'_j}}{\sum{M'}}$, where $\sum{M'_j}$ is the summation of all pixel values in the $j$-th grid cell while $\sum{M'}$ is the sum of all pixels. (3) If the probability of being focused is larger than the threshold $\frac{1}{n \cdot m}$, the entry of this region will be set to $1$, otherwise to $0$. Fig. \ref{fig:grid vector} shows an example of this procedure. 

Given the grid setting $n$ and $m$, the encoded feature $v=\mathcal{E}(X)$ and the grid-vector $y$ transformed from the ground-truth saliency map $M$, we train the gaze prediction module $\mathcal{G}(\cdot)$ using the binary cross-entropy loss:
\begin{equation}
\label{eq:bce loss}
    L(\hat{y},y)= - \frac{1}{K}\sum_{i=1}^{K}y_i \cdot log(\hat y_i) + (1-y_i) \cdot (1-log(\hat y_i))
\end{equation}
where $\hat{y} = \mathcal{G}(v)$ and $K = n\cdot m$ represents the number of grid cells. 

To get a 2D saliency map, we conduct $\hat{M} = \mathcal{T}(\hat{y})$. More specifically, each entry in $\hat{y}$ represents a grid cell in the 2D map (see \cref{fig:grid vector}) and we fill each grid with its entry value. The size of each grid cell is $\frac{H}{n} \times \frac{W}{m}$, therefore a 2D matrix in the size of $n \times m$ is constructed. Then we apply a Gaussian blur and softmax to smooth the 2D matrix and use it as the predicted saliency map $\hat{M}$.
% we re-arranged the output as grid, used bilinear interpolation for resizing to $36 \times 64$ and applied a gaussian blur ($\sigma = 5$) and softmax.
The upper branch in \cref{fig:detailed framework} shows the procedure of predicting a grid-based saliency map.

\label{sec:framework}

\paragraph{Attention-based Object Detection Algorithm.} An object detector $\mathcal{D}(\cdot)$ takes $v$ as input and predicts all objects' information $\mathbf{O}$: the classes and bounding box. Our feature encoder $\mathcal{E}(\cdot)$ together with $\mathcal{D}(\cdot)$ form an entire object detection network. To train a good object detector, a large image dataset with densely annotated (bounding boxes and classes) information is required. Since there are some well-trained publicly available object detection models, e.g., YOLOv5 \cite{glenn_jocher_2021_4679653}, we use their pretrained parameters in our $\mathcal{E}(\cdot)$ and $\mathcal{D}(\cdot)$. More details about the architecture design will be discussed in the next section. Please note that we do not require extra training on $\mathcal{E}(\cdot)$ or $\mathcal{D}(\cdot)$, which makes our whole framework fast to train. Given all objects' information $\mathbf{O}$ and a saliency map $\hat{M}$, the attention-based object detection operation $\bigotimes$ works as follows: for each object $\mathbf{o_i} \in \mathbf{O}$, we use the \yao{maximum} pixel value inside its bounding box area on $\hat{M}$ as the probability of being focused for $\mathbf{o_i}$. A threshold $Th$ for the probability can be set to detect whether $\mathbf{o_i}$ is \yao{focused on} by drivers. $Th$ can be chosen by users according to their requirements for different metrics, such as precision or recall. A separate discussion regarding the effect of $Th$ can be found in \cref{sec:results}.

%we map object bounding boxes to the saleincy map. An object is detected as being focused by drivers, if at least one pixel within the bounding box reaches the threshold $Th$.

\subsection{Model Details}
\label{sec:implementation}
We use three pretrained object detection networks as our feature encoder $\mathcal{E}(\cdot)$, i.e., YOLOv5 \cite{glenn_jocher_2021_4679653}, Gaussian YOLOv3 \cite{choi2019gaussian} and CenterTrack \cite{zhou2020tracking}, to validate the efficiency and adaptability of our gaze prediction. Specifically, we deploy the layers in the YOLOv5 framework (size small, release v5.0) before the last CSP-Bottleneck (Cross Stage Partial \cite{wang2020cspnet}) layer within the neck (PANet \cite{liu2018path}). Meanwhile, we use the remaining part of the model (i.e., the detector layer) as the object detector $\mathcal{D}(\cdot)$. Similarly, we use the partial network of YOLOv3 (first 81 layers) as $\mathcal{E}(\cdot)$, and use the ``keypoint heatmaps'' for every class of CenterTrack \cite{zhou2020tracking}. \cref{tab:architecture} lists the concrete dimension of extracted $v$. Furthermore, this table also presents the dimension of the output after each layer in the gaze prediction module. The convolutional layer with the kernel size $1 \times 1$ shrinks the input channels to 16 when using YOLO backbones, while to one channel when the CenterTrack features are used. To reduce the computational burden for the dense layer, an average pooling layer is deployed to reduce the width and height of the feature maps. Before being put into the dense layer, all the features are reshaped to vectors. The dense layer followed by the sigmoid activation function outputs the $\hat{y} \in \mathcal{R}^{n \cdot m}$.

% We shrank the input channels again with an $1 \times 1$ kernel convolutional layer to 16 and halved the feature maps dimension with an average pooling layer to $7 \times 7$. With CenterTrack we used the output feature maps of size $80 \times 128 \times 128$. We cropped padded zeros at top and bottom to the height of 72, shrank the channel dimension with an $1 \times 1$ kernel convolutional layer to one and shrank the feature size with an average pooling layer to $18 \times 32$.  

% For object detection we use the network YOLO v5 version 5.0 short \cite{glenn_jocher_2021_4679653} and extract the feature maps of size $512 \times 12 \times 20$ after Layer 22, before the latest BottleNeckCSP, to use the same backbone in the gaze prediction network. The gaze prediction network adds only a convolutional layer with kernel $1 \times 1$ to shrink the channels to 16, an average pooling layer to shrink the feature map size to $6 \times 10$ and a dense layer with sigmoid activation function. The dense layer output dimension is given by the number of grid cells.

\begin{table}[h]
\caption{Network architecture details when using different object detectors. Column ``Feature Encoder'' shows the used backbone for extracting feature $v$ and the dimension of $v$. Column ``Gaze Prediction'' demonstrates the dimension of output after each layer.}
\label{tab:architecture}
\resizebox{.9\linewidth}{!}{
\begin{tabular}{cc|ccc}
\hlinewd{1pt}
\multicolumn{2}{c|}{Feature Encoder $\mathcal{E}(\cdot)$} & \multicolumn{3}{c}{Gaze Prediction $\mathcal{G}(\cdot)$}                                        \\ \hline
\multicolumn{1}{c|}{Backbone}    & $v$ & \multicolumn{1}{c|}{Conv} & \multicolumn{1}{c|}{Avg Pooling} & Dense Layer \\ \hlinewd{1pt}
\multicolumn{1}{c|}{YOLOv5 \cite{glenn_jocher_2021_4679653}}      &  $512 \times 12 \times 20$  & \multicolumn{1}{c|}{$16 \times 12 \times 20$}     & \multicolumn{1}{c|}{$16 \times 6 \times 10$}            &      number of grid cells         \\ \hline
\multicolumn{1}{c|}{Gaussian YOLOv3 \cite{choi2019gaussian} }      & $1024 \times 13 \times 13$   & \multicolumn{1}{c|}{$16 \times 13 \times 13$}     & \multicolumn{1}{c|}{{$16 \times 7 \times 7$} }            &      number of grid cells           \\ \hline
\multicolumn{1}{c|}{CenterTrack \cite{zhou2020tracking}} & {$80 \times 72 \times 128$}   & \multicolumn{1}{c|}{$1 \times 72 \times 128$}     & \multicolumn{1}{c|}{$1 \times 18 \times 32$}            &     number of grid cells            \\ \hlinewd{1pt}
\end{tabular}
}
\end{table}

\section{Experimental Results}
\label{sec:results}
% \textcolor{blue}{ I will work on this section. We first give implementation details here, e.g. how many epochs we train, lr, optimizer, dataset details.}
In this section, we first introduce \naemi{experimental implementation including analysis of the datasets BDD-A and DR(eye)VE, evaluation metrics and the details of how we train our proposed gaze prediction module on the BDD-A dataset. After the implementation details,  we show and discuss the evaluation results of our whole framework on attention prediction as well as attention-based object detection compared to other state-of-the-art driver attention prediction networks.} To further validate the effectiveness of our network, we tested and evaluated our framework on several videos from the DR(eye)VE dataset \cite{alletto2016dr}.

% We trained and evaluated our framework on the BDD-A dataset \cite{xia2018predicting} using its default training-validation-test split. Besides the BDD-A dataset, we also conducted evaluation on the DR(eye)VE dataset \cite{alletto2016dr} without any training. In this section, we discuss the results of using both datasets. Training details can be found in Section \ref{sec:bdda}.

\subsection{\yao{Implementation Details}}
\subsubsection{\naemi{Datasets}}
\paragraph{\yao{BDD-A}}
The BDD-A dataset  \cite{xia2018predicting} includes a total of 1426 videos, each is about ten seconds in length. Videos were recorded in busy areas with many objects on the roads. There are 926 videos in the training set, 200 in the validation set and 300 in the test set. We extracted three frames per second and after excluding invalid gaze maps, the training set included 30158 frames, the validation set included 6695 frames and the test set 9831. \cref{tab:bdda analysis} shows the statistics of the ground-truth ``focused on'' objects on the test set. In each image frame, there are on average 7.99 cars detected (denoted as ``Total''), whereas 3.39 cars of those attract the driver's attention (denoted as ``Focused''). 0.94 traffic lights can be detected in each frame, but only 0.18 traffic lights are noticed by the driver. This is due to the fact that drivers mainly attend to traffic lights that are relative to their driving direction. In total, there are 10.53 objects and approximately 40\% (4.21 objects) fall within the driver's focus. Therefore, to accurately detect these focused objects is challenging.

\begin{table}[h!]
\caption{Traffic-related class analysis on BDD-A test set: The values in the table show the average number of objects in one video frame. ``Total'' means detected objects while ``focused'' means attended objects by the human driver. ``-'' refers to a number smaller than 0.001. ``Sum'' includes also non-traffic objects.}
%(The sum includes all 80 classes of Coco dataset.)
\label{tab:bdda analysis}
\resizebox{.7\textwidth}{!}{% <------ Don't forget this %
\begin{tabular}{|l|l|l|l|l|l|l|}
\hline
\textbf{Object}   & \textbf{Person}        & \textbf{Bicycle}      & \textbf{Car}       & \textbf{Motorcycle}    & \textbf{Bus}   & \textbf{Truck}                     \\ \hline
Total    & 0.78         & 0.03       & 7.99 & 0.03          & 0.18   & 0.48                     \\ \hline
Focused & 0.24         & 0.02          & 3.39     & 0.01           & 0.11   & 0.25                      \\ \hline
\textbf{Object}   & \textbf{Traffic light} & \textbf{Fire Hydrant} & \textbf{Stop Sign} & \textbf{Parking Meter} & \textbf{Bench} & \textit{\textbf{Sum}} \\ \hline
Total   & 0.94         & 0.02          & 0.05       & 0.004            & 0.002    & 10.53                           \\ \hline
Focused & 0.18         & 0.002          & 0.008        & -             & -     &
4.21 \\ \hline
\end{tabular}
}
\end{table}

\paragraph{\yao{DR(eye)VE}}
The DR(eye)VE dataset \cite{alletto2016dr} contains 74 videos. We used five videos (randomly chosen) from the test set (video 66, 67, 68, 70 and 72), which cover different times, drivers, landscapes and weather conditions. Each video is 5 minutes long and the FPS (frames per second) is 25, resulting in 7500 frames for each video. After removing frames with invalid gaze map records, our test set includes 37270 frames in total. We run a pretrained YOLOv5 network on all five videos and obtained the results shown in Table \ref{tab:dreyeve analysis}. Compared to the BDD-A dataset in Table \ref{tab:bdda analysis}, DR(eye)VE incorporates a relatively monotonous environment with fewer objects on the road. On average, there are 3.24 objects in every frame image. 39\% of the objects are attended by drivers, which is similar to the BDD-A dataset.

\begin{table}[h]
\caption{Traffic-related class analysis on DR(eye)VE dataset (test set): The value is the average number of objects in each video frame. ``Total'' means detected objects while ``focused'' means attended objects by the human driver. ``-'' refers to the number smaller than 0.001. ``Sum'' includes also non-traffic objects.} 
%(The sum includes all 80 classes of Coco dataset.)}
\label{tab:dreyeve analysis}
\resizebox{.7\linewidth}{!}{% <------ Don't forget this %
\begin{tabular}{|l|l|l|l|l|l|l|}
\hline
\textbf{Object}   & \textbf{Person}        & \textbf{Bicycle}      & \textbf{Car}       & \textbf{Motorcycle}    & \textbf{Bus}   & \textbf{Truck}                     \\ \hline
Total    & 0.07        & 0.009         & 2.35  & 0.003   & 0.026  & 0.09                      \\ \hline
Focused & 0.02          & 0.004       & 1.06   & -         & 0.01   & 0.04                   \\ \hline
\textbf{Object}   & \textbf{Traffic light} & \textbf{Fire Hydrant} & \textbf{Stop Sign} & \textbf{Parking Meter} & \textbf{Bench} & \textit{\textbf{Sum}} \\ \hline
Total    & 0.46       & -          & 0.02      & 0.005            & 0.003    & 3.24                         \\ \hline
Focused & 0.07        & -           & 0.002      & 0.003             & -     &
1.26 \\ \hline
\end{tabular}
}
\end{table}

\subsubsection{\naemi{Evaluation Metrics}}\hfill \break
We evaluated the models from \naemi{three} perspectives: object detection (object-level), saliency map generation (pixel-level) \naemi{and resource costs}. To compare the quality of generated gaze maps, we used the Kullback–Leibler divergence ($D_{KL}$) and \naemi{Pearson's} Correlation Coefficient ($CC$) metrics as in previous works \cite{xia2018predicting,dreyeve2018,pal2020looking}. We resized the predicted and ground-truth saliency maps to $36\times64$ \naemi{keeping the original width and height ratio following the setting of Xia et al. \cite{xia2018predicting}. Since saliency maps predicted by different models were in different sizes, we scaled them to the same size ($36\times64$) as suggested by Xia et al. \cite{xia2018predicting} to fairly compare them.}
For the object detection evaluation, we first decided the ground-truth ``focused'' objects by running our attention-based object detection on all the objects (detected by the YOLOv5 model) and the ground-truth gaze saliency maps, ${M} \bigotimes \mathbf{O}$, i.e., used the maximal value inside the object (bounding) area as the probability. If that probability was larger than 15\%, this object was recognized as the ``focused on'' object. The 15\% was chosen empirically to filter out the objects that were less possible than a random selection (averagely ten objects in one frame shown in \cref{tab:bdda analysis}). For the evaluation, we regarded each object as a binary classification task: the object was focused by the driver or not. \naemi{The evaluation metrics used here were Area Under ROC Curve ($AUC$), precision, recall, $F_1$ score and accuracy. Except for $AUC$, all the metrics require a threshold $Th$, which will be discussed in \cref{sec:bdda}. Finally, to quantitatively measure and compare the computational costs of our models, we considered the number of trainable parameters and the number of floating point operations (GFLOPs) of the networks.}

\subsubsection{\naemi{Training Details}}\hfill \break
All experiments were conducted on \naemi{one NVIDIA CUDA RTX A4000 GPU. The proposed gaze prediction module was trained for 40 epochs on the BDD-A training set} using the Adam optimizer \cite{kingma2015adam} and validated on the validation set. The learning rate started from 0.01 and decayed with a factor of 0.1 after every 10 epochs. The feature encoder and the object detector were pretrained\footnote{Pretrained parameters for YOLOv5 can be found at \url{https://github.com/ultralytics/yolov5}; for YOLOv3 at \url{https://github.com/motokimura/PyTorch_Gaussian_YOLOv3} and for CenterTrack at \url{https://github.com/xingyizhou/CenterTrack}.} and we did not require further fine-tuning for the object detection. %\yao{\textit{@Naemi, please also add the hardware information here.}}

\subsection{Results on BDD-A}
\label{sec:bdda}
\subsubsection{\naemi{Quantitative Results}}
\paragraph{\naemi{Different Grids}} We first conducted experiments on different grid settings in the gaze prediction module: from 2$\times$2 ($n=m=2$) to 32$\times$32 ($n=m=32$) increasing by a factor of 2. We used YOLOv5 as our backbone for all grid settings here. The evaluation between different grids is shown in \cref{tab:different grids}. ``Pixel-level'' refers to the evaluation of the saliency map using $D_{KL}$ and $CC$ metrics. ``Object-level'' refers to results of attention-based object detection. We set the threshold $Th$ for detecting attended regions to 0.5 to compare the performance between different settings fairly. This evaluation shows that the performance increases when the grids become finer. Nevertheless, we can see that the advantage of 32$\times$32 grids over 16$\times$16 grids is not significant and the $AUC$ is almost equal. To save computational costs, we chose the 16$\times$16 grids as our model setting for all further experiments. 

\begin{minipage}[c]{0.6\linewidth}
\centering
%\begin{table}[h!]
\captionof{table}{Comparison of using different grid settings on object- and pixel-level performance ($Th$=0.5). For all metrics except $D_{KL}$, a higher value indicates the better performance. The best result is marked in bold.}
\label{tab:different grids}
\resizebox{\linewidth}{!}{% <------ Don't forget this %
\begin{tabular}{c|c|c|c|c|c|c|c|}
\cline{2-8}
& \multicolumn{5}{c|}{\textbf{Object-level}}                   & \multicolumn{2}{c|}{\textbf{Pixel-level}} \\ \cline{2-8} 
& \textit{AUC} & \textit{Prec (\%)} & \textit{Recall (\%)} & \textit{$F_1$ (\%)} & \textit{Acc (\%)} & \textit{$D_{KL}$}         & \textit{CC}         \\ \hline
\multicolumn{1}{|c|}{\textbf{2$\times$2}}   & 0.58   &    43.86          &     88.97       &  58.75    &   50.05    &     2.35         &      0.18         \\ \hline
\multicolumn{1}{|c|}{\textbf{4$\times$4}}   &  0.76    &    52.43      &    \textbf{91.50}       &  66.66     &   63.40  &     1.61           &        0.41        \\ \hline
\multicolumn{1}{|c|}{\textbf{8$\times$8}}   &   0.84   &     57.87       &     89.16      &   70.18   &   69.71    &    1.27  &   0.55             \\ \hline
\multicolumn{1}{|c|}{\textbf{16$\times$16}} & \textbf{0.85} & 71.98 & 73.31  & \textbf{72.64}   &  77.92 &  1.15 &  0.60      \\ \hline
\multicolumn{1}{|c|}{\textbf{32$\times$32}} & \textbf{0.85}     &   \textbf{75.47}    &    68.79     & 71.97    &  \textbf{78.58}   &   \textbf{1.13}        &   \textbf{0.62}            \\ \hline
\end{tabular}
}
\end{minipage}
%\end{table}
\hfill
\begin{minipage}[c]{0.35\linewidth}
\centering
%\begin{table}[h!]
\captionof{table}{Comparison of different $Th$ using 16$\times$16 grids on attention-based object detection. Results are shown in \% and for all metrics, a higher value indicates better performance. The best result is marked in bold.}
%\caption{Comparison of different Th for YOLOv5 grid 16$\times$16 models on BDD-A test set. For all metrics the higher value represents better performance.}
\label{tab:different Th on bdda}
\resizebox{\linewidth}{!}{% <------ Don't forget this %
\begin{tabular}{c|c|c|c|c|}
\cline{2-5}
& \textbf{Prec} & \textbf{Recall} & $\mathbf{F_1}$ & \textbf{Acc }       \\ \hline
\multicolumn{1}{|c|}{\textbf{0.3}}   & 63.76          &    \textbf{83.33}       &  72.24    &   74.39         \\ \hline
\multicolumn{1}{|c|}{\textbf{0.4}}   & 68.11          &     78.36       &  \textbf{72.88}    &   76.68         \\ \hline
\multicolumn{1}{|c|}{\textbf{0.5}}   & 71.98          &     73.31       &  72.64    &   77.92         \\ \hline
\multicolumn{1}{|c|}{\textbf{0.6}}   & 75.81          &     68.09       &  71.74    &   \textbf{78.55}         \\ \hline
\multicolumn{1}{|c|}{\textbf{0.7}}   & \textbf{79.61}          &     62.04       &  69.73    &   78.47         \\ \hline
\end{tabular}
}
\end{minipage}
%\end{table}
\\

\paragraph{\naemi{Different Thresholds}}  The effect of different $Th$ on attention-based object detection is listed in \cref{tab:different Th on bdda}. Our results show that a lower $Th$ yields better performance on the recall score, while a higher $Th$ improves the precision score. The best $F_1$ score is achieved when $Th$ is equal to 0.4, and for the best accuracy $Th$ is set to 0.6. When setting $Th$ to 0.5, we obtain relatively good performance in $F_1$ (72.64\%) and in the accuracy (77.92\%). $Th$ is a hyperparameter that users can decide according to their requirements for the applications. For example, if high precision is preferred, $Th$ can be set to a higher value.
%We have also tried grids $16 \times 32$ ($AUC = 0.85, \ KL 1.16, \ CC=0.61$) and $32 \times 16$ ($AUC = 0.85, \ KL 1.15, \ CC=0.61$), but this has not improved the $AUC$ value either. 

%The grid size determines the number of classes of the critical region prediction network. For example, if the grid is 4$\times$4, the number of classes is 16. 
% We conducted experiments on additional variants of grid $16 \times 32$ ($AUC = 0.85, \  KL = 1.16, \ CC = 0.61$) and $32 \times 16$ ($AUC = 0.85, \  KL = 1.15, \ CC = 0.61$), but we cannot achieve better AUC and therefore we chose the lightest grid  $16 \times 16$ as our final model. In Table \ref{tab:different Th}, we show the object-level metrics for different treshholds on $16 \times 16$-grid. Since $Th = 0.5$ is good in $F1$ and $Acc$, we will use this one in further evaluations.

\paragraph{\naemi{Comparison with other Models}} 
We compared our \naemi{three} proposed model\naemi{s based on YOLOv5, Gaussian YOLOv3 and CenterTrack} with \naemi{four existing saliency models}: BDD-A \cite{xia2018predicting}, DR(eye)VE \cite{dreyeve2018}, ML-Net \cite{cornia2016deep} and PiCANet \cite{liu2018picanet}\footnote{All models were trained on the BDD-A training set. Trained parameters of the BDD-A model were downloaded from \url{https://github.com/pascalxia/driver_attention_prediction} and the rest were from \url{https://sites.google.com/eng.ucsd.edu/sage-net}.}. 
% Furthermore, we tried our approach with two other object detection networks, Gaussian YOLOv3 and CenterTrack\footnote{Trained Gaussian YOLO v3 model is downloaded from \url{https://github.com/motokimura/PyTorch_Gaussian_YOLOv3}, trained CenterTrack model from \url{https://github.com/xingyizhou/CenterTrack}.}. For better comparison, we also choose grid $16 \times 16$ and evaluate the object-level with the detected objects of YOLO v5. All models were trained on the BDD-A training set. 
We examined the performance from three perspectives: object detection, gaze saliency map generation and resource cost. 
For the object detection, we used the same object detector (YOLOv5) to detect all objects in images, then run our attention-based object detection algorithm $\bigotimes$ based on generated saliency maps from each model. \naemi{The ``Baseline'' refers to the average BDD-A training set saliency map as illustrated in \cref{fig:saliency prediction} (b). For a fair comparison of the $Th$-dependent object-level scores precision, recall, $F_1$ and accuracy, we computed for each model the threshold $Th$, which gives the best ratio of the true positive rate (TPR) and the false positive rate (FPR). Specifically, we created for each model the ROC curve (Receiver Operating Characteristic) on the BDD-A test set and determined the $Th$, which corresponds to the point on the curve with the smallest distance to (0,1): $argmax(\sqrt{TPR \cdot (1-FPR)})$. The ROC curves and the values of $Th$ for each model can be found in \cref{sec:appendix6}. \cref{tab:comparison in bdda} shows the results of our comparison with the different models}. (More results of using other $Th$ can be found in \cref{sec:appendix1}.) 

%The object detection and gaze map show the performance of models in object-level and pixel-level, respectively.
% Since all models use an object detection network, i.e., YOLOv5, we only mention additional costs as resource costs. 
%shows the evaluation and figure \ref{fig:images} shows an example prediction for all models.
%The baseline model shows that only predicting center parts as driver attention regions is not, neither at the pixel-level nor at the object-level, precise enough. 
The AUC scores show that our two YOLO models can compete on object level with the other models, even though PiCANet performs slightly better. Although our models were not trained for pixel-level saliency map generation, the $D_{KL}$ and $CC$ values show that our YOLOv5 based model with $D_{KL}$ of 1.15 and $CC$ of 0.60 is even on pixel-level comparable to the other models (under our experiment settings). In object detection, our two YOLO-based models achieve 0.85 in the $AUC$, which is slightly inferior to PiCANet of 0.86. Nevertheless, they have better performance in $F_1$ and accuracy scores than other models.

Moreover, our gaze prediction model shares the backbone (feature encoder) with the object detection network and requires mainly one extra dense layer, which results in less computational costs. For instance, our YOLOv5 based model requires 7.52M parameters in total and only 0.25M from them are extra parameters for the gaze prediction, which results in the same computational cost as a YOLOv5 network (17.0 GFLOPs). In general, the advantage of our framework is that the gaze prediction almost does not need any extra computational costs or parameters than the object detection needs. Other models need an extra object detection network to get the attention-based objects in their current model architectures. Nevertheless, we list the needed resources of each model only for the saliency prediction in \cref{tab:comparison in bdda} for a fair comparison. To achieve a similar object detection performance, for example, DR(eye)VE needs 13.52M parameters and 92.30 GFLOPs to compute only saliency maps, which are more than our YOLOv5 framework requires for the object detection task and saliency map prediction together.
%to compute saliency maps first, which are more than our YOLOv5 framework requires.
%Please note that for our models the listed resource is for the object detection and saliency prediction in total., requiring the additional computation of YOLOv5 (7.27M parameters and 16.5 GFLOPs) for object detection. 
% This is why our models have such low resource costs and are very competitive in practical implementation.

\begin{table}[t]
\caption{Comparison with other gaze models on the BDD-A dataset. On object-level, all models are evaluated with detected objects of YOLOv5. Our three models use 16$\times$16 grids. Pixel-level values in brackets are the results reported from the original work~\cite{xia2018predicting, pal2020looking}. * indicates that the backbone is pretrained on COCO \cite{lin2014microsoft}, $\dagger$ on ImageNet \cite{deng2009imagenet} and $\ddagger$ on UCF101 \cite{soomro2012ucf101}. The resource required for the gaze prediction is listed in the last column.}
%It has also to be mentioned that the other approaches use older architectures as backbone.
\label{tab:comparison in bdda}
\resizebox{\linewidth}{!}{% <------ Don't forget this %
\begin{tabular}{c|c|c|c|c|c|c|c|c|c|}
\cline{2-10}
& \multicolumn{5}{c|}{\textbf{Object-level}} 
& \multicolumn{2}{c|}{\textbf{Pixel-level}}  &
\multicolumn{2}{c|}{\textbf{Resource}} \\ \cline{2-10}
& \textit{AUC} &  \textit{Prec.  (\%)} & \textit{Recall  (\%)} & \textit{$F_1$  (\%)} & \textit{Acc  (\%)} & \textit{$D_{KL}$} & \textit{CC}  &\textit{Param.(M)}  & \textit{GFLOPs}    \\ \hline
\multicolumn{1}{|c|}{\textbf{Baseline}}  &   0.82  & 66.10 &  74.22  &  69.92 & 74.47 & 1.51  & 0.47 & 0.0 & 0.0  \\ \hline
\multicolumn{1}{|c|}{\textbf{BDD-A} \cite{xia2018predicting} $^\dagger$} & 0.82 & 66.00             & 74.33       & 69.92    & 74.43      &    1.52 (1.24)   &  0.57 (0.59)    & 3.75 & 21.18     \\ \hline
\multicolumn{1}{|c|}{\textbf{DR(eye)VE} \cite{dreyeve2018} $^\ddagger$} & 0.85 & 70.04            & 74.94          & 72.41      & 77.16         &    1.82 (1.28)    &  0.57  (0.58)    & 13.52 &    92.30 \\ \hline
\multicolumn{1}{|c|}{\textbf{ML-Net} \cite{cornia2016deep}$^\dagger$} & 0.84 & 70.48 & 73.75          & 72.08       & 77.15         &    1.47 (1.10) &     0.60 (0.64)   &15.45 & 630.38    \\ \hline
\multicolumn{1}{|c|}{\textbf{PiCANet} \cite{liu2018picanet}$^\dagger$} & 0.86 & 70.23           & 77.67        & 73.76      & 77.91      &    1.69 (1.11)        &      0.50 (0.64)     & 47.22 & 108.08 \\ \hline
\multicolumn{1}{|c|}{\textbf{Ours (CenterTrack)}*}  & 0.83  & 68.93 & 72.83  &  70.83   & 76.01  & 1.32  & 0.56   & 19.97 & 28.57     \\ \hline
\multicolumn{1}{|c|}{\textbf{Ours (YOLOv3)}*} & 0.85  & 70.25  & 74.72 & 72.41  &  77.24 & 1.20  & 0.59  & 62.18  &  33.06 \\ \hline
\multicolumn{1}{|c|}{\textbf{Ours (YOLOv5)}*} & 0.85 & 70.54 & 75.30   & 72.84   & 77.55 &  1.15 &  0.60  & 7.52 & 17.0    \\ \hline
\end{tabular}
}
\end{table}

\subsubsection{\naemi{Qualitative Results}}
\begin{figure}[t]
\captionsetup[subfigure]{aboveskip=-.6pt}
\centering
%\resizebox{\linewidth}{!}{% <------ Don't forget this %
\begin{subfigure}{.18\linewidth}
    \centering
    \includegraphics[width=\linewidth]{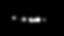}
    \caption{Ground-truth}
\end{subfigure}
\hfill
\begin{subfigure}{.18\linewidth}
    \centering
       \includegraphics[width=\linewidth]{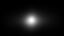}
    \caption{Baseline}
\end{subfigure}
\hfill
\begin{subfigure}{.18\linewidth}
    \centering
       \includegraphics[width=\linewidth]{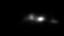}
    \caption{BDD-A}
\end{subfigure}
\hfill
\begin{subfigure}{.18\linewidth}
    \centering
    \includegraphics[width=\linewidth]{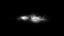}
    \caption{DR(eye)VE}
\end{subfigure}
\hfill
\begin{subfigure}{.18\linewidth}
    \centering
       \includegraphics[width=\linewidth]{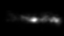}
    \caption{ML-Net}
\end{subfigure}
\\
\begin{subfigure}{.18\linewidth}
    \centering
       \includegraphics[width=\linewidth]{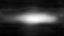}
    \caption{PiCANet}
\end{subfigure}
\hfill
\begin{subfigure}{.18\linewidth}
    \centering
    \includegraphics[width=\linewidth]{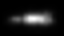}
    \caption{Ours (YOLOv5)}
\end{subfigure}
    \hfill
\begin{subfigure}{.18\linewidth}
    \centering
       \includegraphics[width=\linewidth]{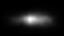}
    \caption{Ours (YOLOv3)}
\end{subfigure}
   \hfill
\begin{subfigure}{.18\linewidth}
    \centering
       \includegraphics[width=\linewidth]{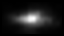}
    \caption{Ours (CenterTr.)}
\end{subfigure}
\caption{Comparison of predicted driver attention saliency maps using different models. (a) Ground-truth driver attention map; (b) The baseline saliency map (center-bias); (c-f) Predictions using models \cite{xia2018predicting,dreyeve2018,cornia2016deep,liu2018picanet}; (g-i) Predictions using our framework with different backbones.}
\label{fig:saliency prediction}
\end{figure}

\hfill \break
We demonstrate the qualitative results of the saliency map prediction using different models in \cref{fig:saliency prediction}. Our framework uses the backbones from YOLOv5, YOLOv3 and CenterTrack. We see that BDD-A, DR(eye)VE and ML-Net provide a more precise and concentrated attention prediction. However, BDD-A and ML-Net highlight a small area at the right side wrongly instead of an area at the left side, while our predictions (g) and (h) focus on the center part as well as the right side. Although our predictions are based on grids, they are less coarse than the ones of PiCANet.

% We ran our model on different BDD-A video frames to demonstrate our predictions with images. 

\begin{figure}[t]
\captionsetup[subfigure]{aboveskip=-.6pt}
\centering
\begin{subfigure}{.3\linewidth}
    \centering
    \includegraphics[width=\linewidth]{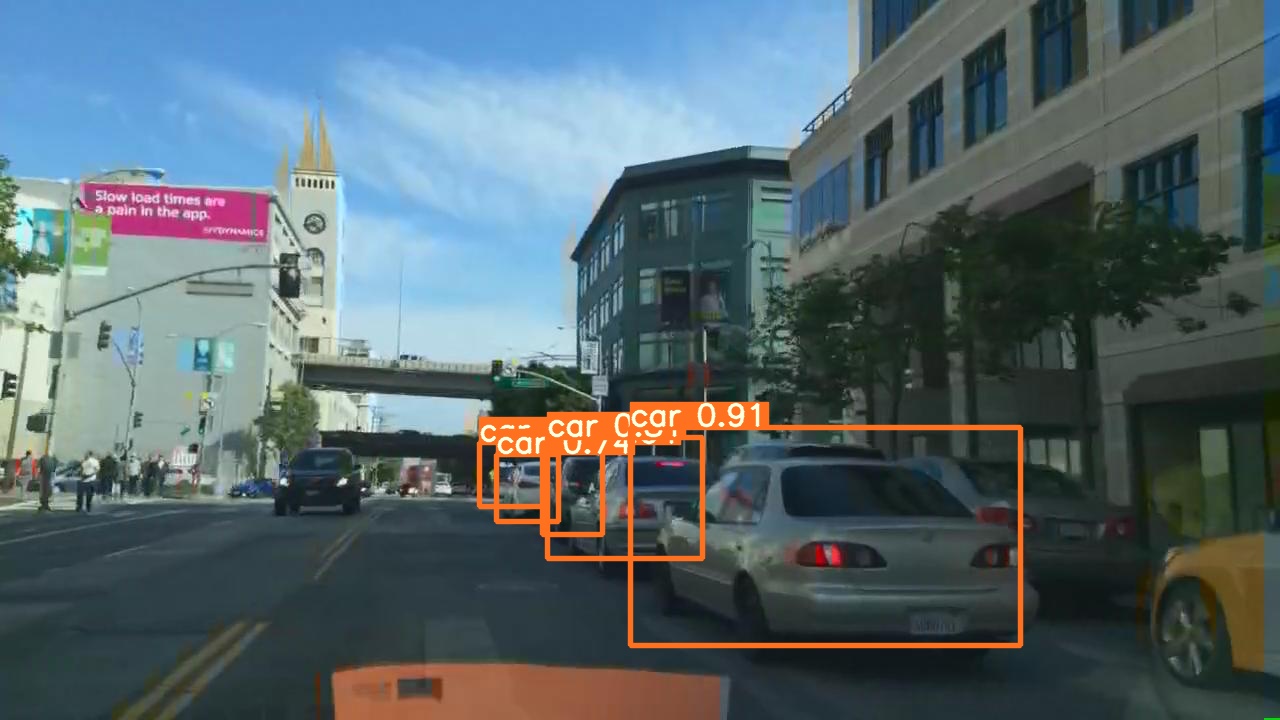}
    \caption{Ground-truth driver attention}
\end{subfigure}
    \hfill
\begin{subfigure}{.3\linewidth}
    \centering
       \includegraphics[width=\linewidth]{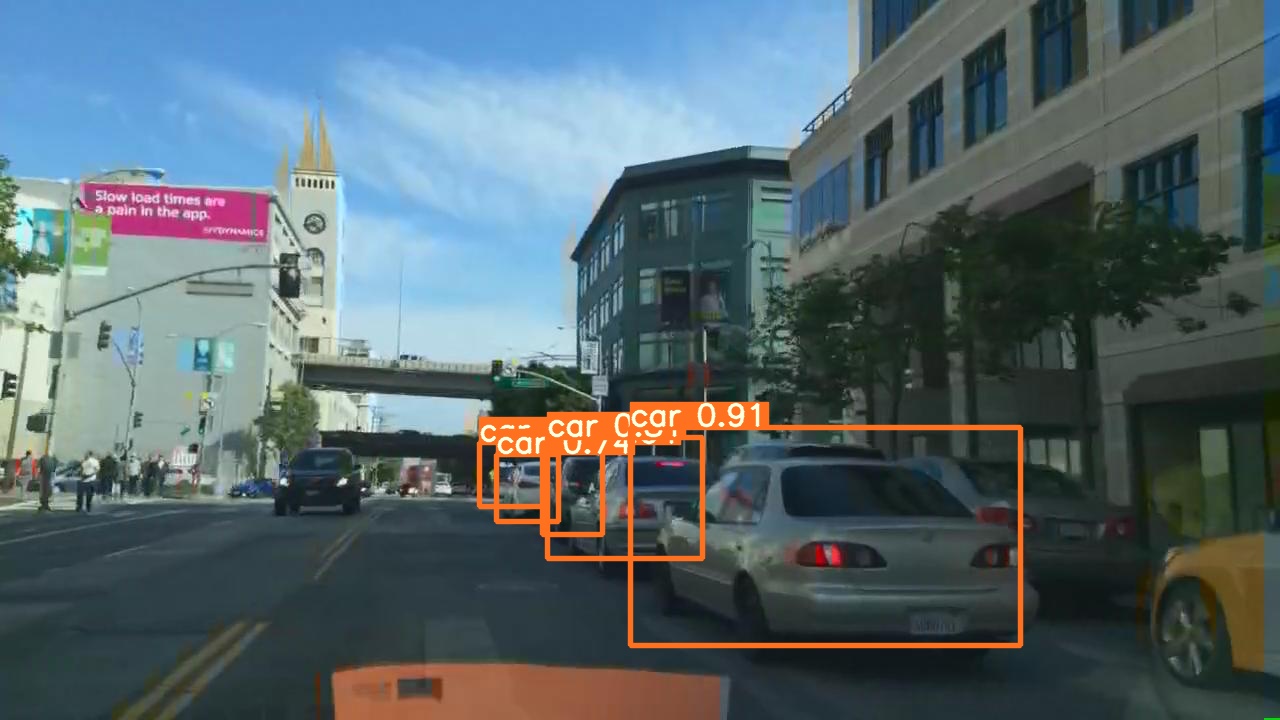}
    \caption{Ours (YOLOv5)}
\end{subfigure}
   \hfill
\begin{subfigure}{.3\linewidth}
    \centering
       \includegraphics[width=\linewidth]{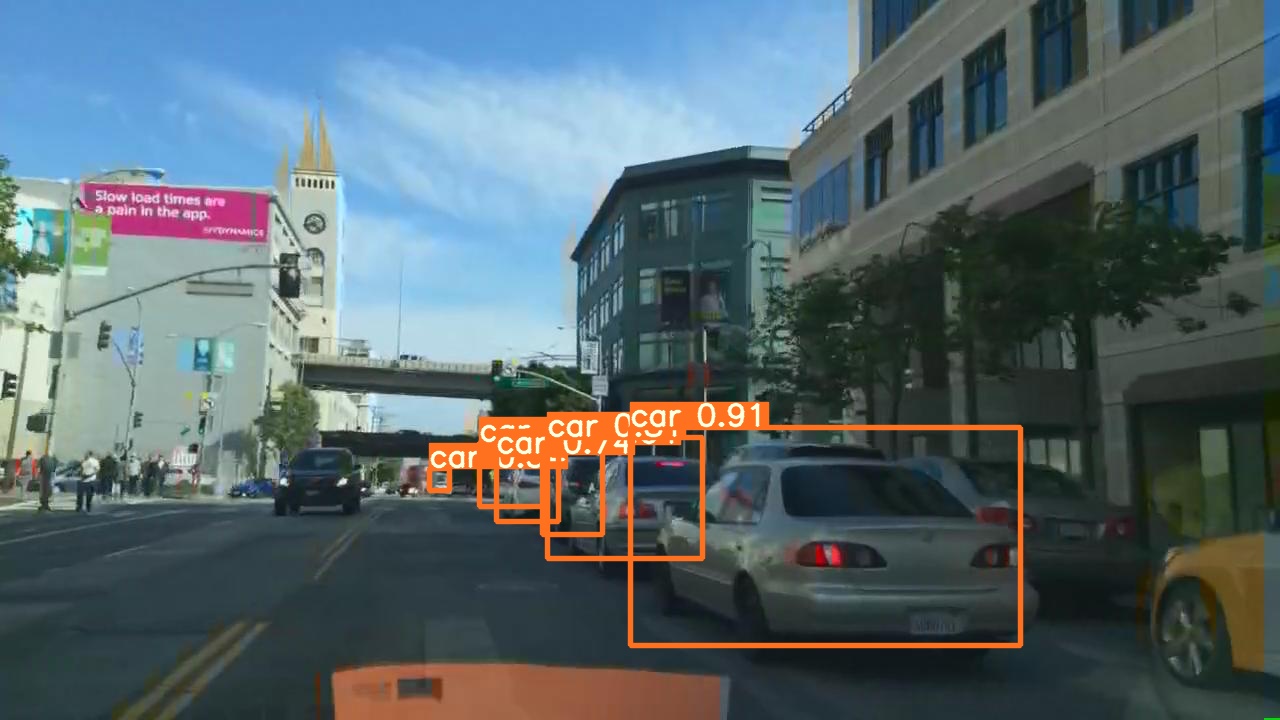}
    \caption{Ours (YOLOv3)}
\end{subfigure}

%\bigskip
\begin{subfigure}{.3\linewidth}
    \centering
    \includegraphics[width=\linewidth]{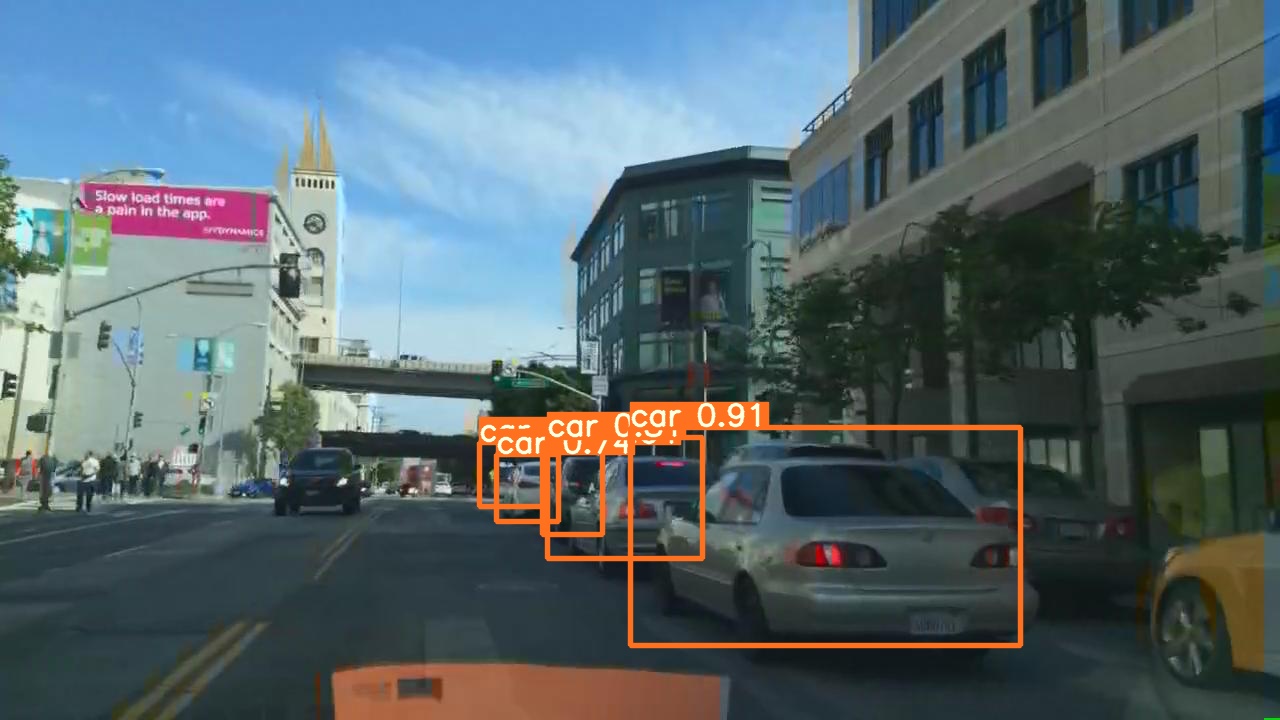}
    \caption{Ours (CenterTrack)}
\end{subfigure}
    \hfill
\begin{subfigure}{.3\linewidth}
    \centering
       \includegraphics[width=\linewidth]{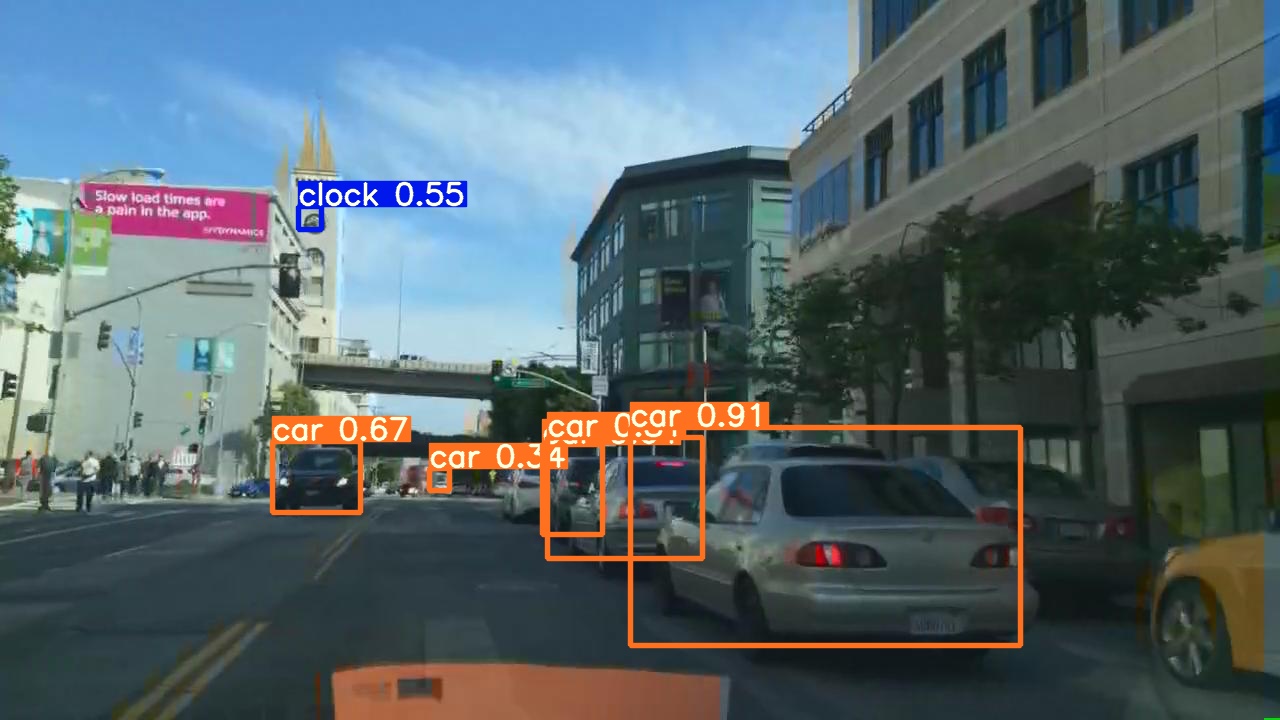}
    \caption{BDD-A}
\end{subfigure}
   \hfill
\begin{subfigure}{.3\linewidth}
    \centering
       \includegraphics[width=\linewidth]{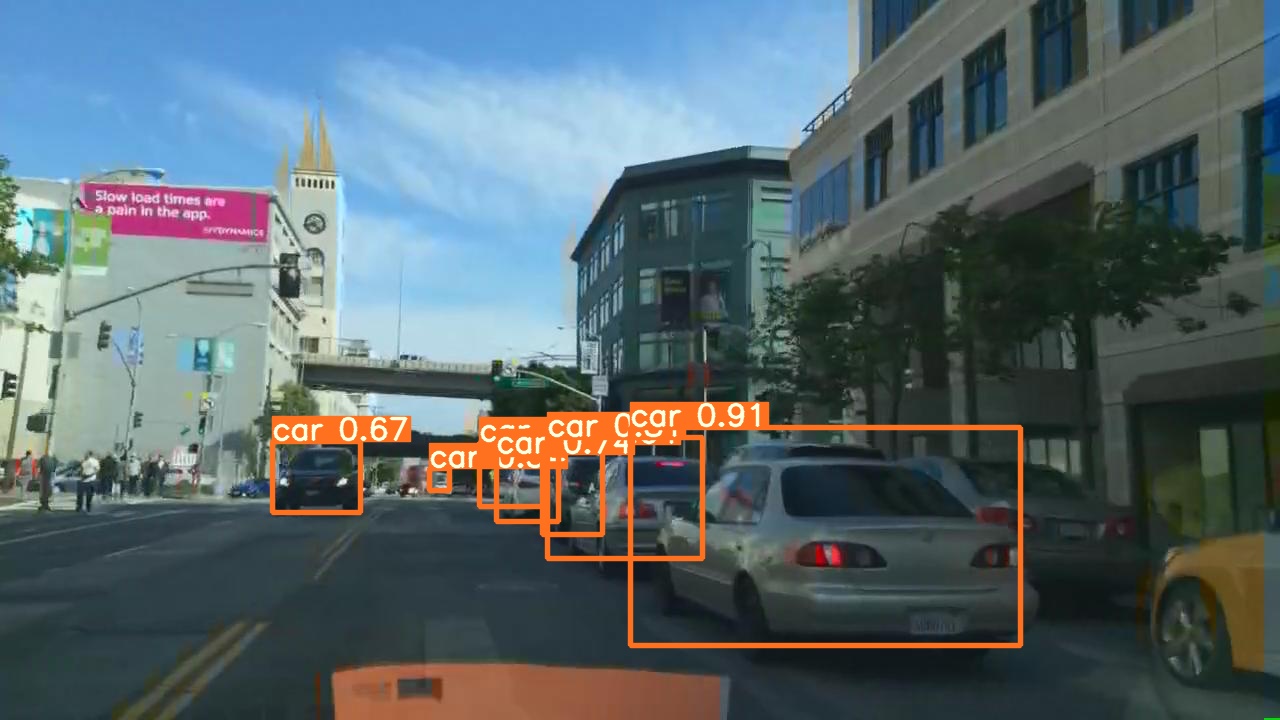}
    \caption{DR(eye)VE}
\end{subfigure}

%\bigskip
\begin{subfigure}{.3\linewidth}
    \centering
       \includegraphics[width=\linewidth]{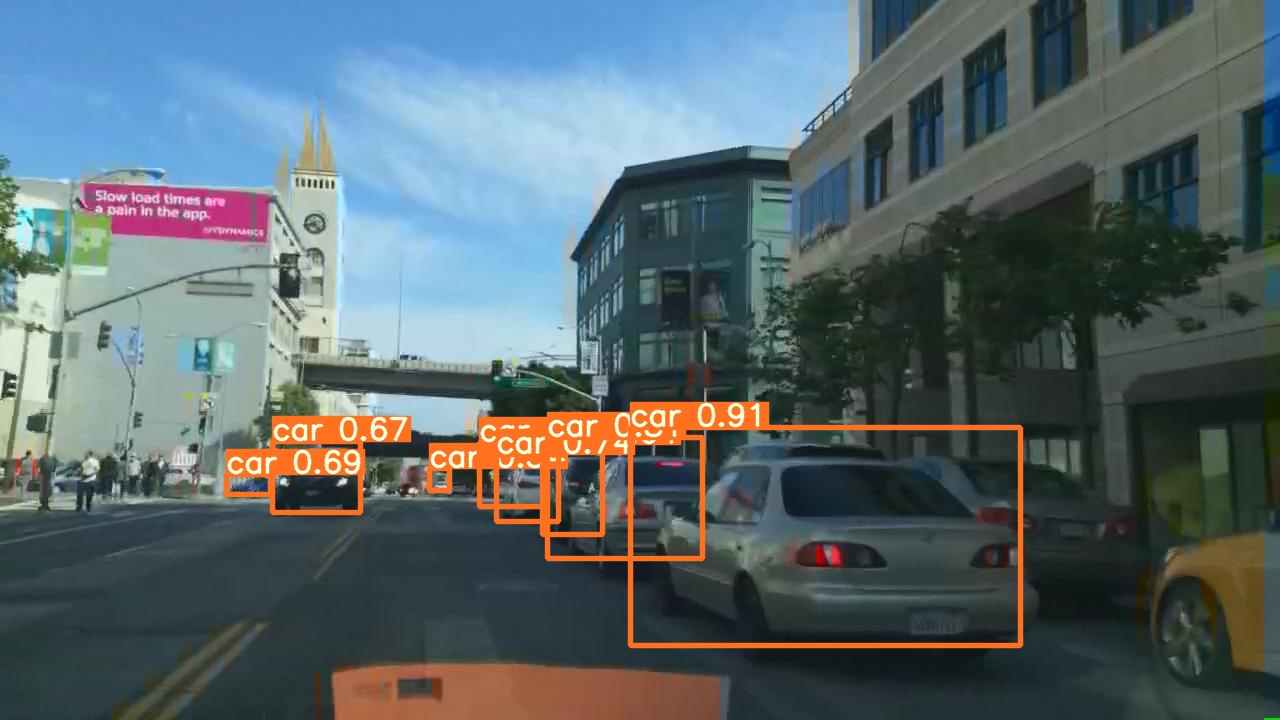}
    \caption{ML-Net}
\end{subfigure}
   \hfill
   \begin{subfigure}{.3\linewidth}
    \centering
    \includegraphics[width=\linewidth]{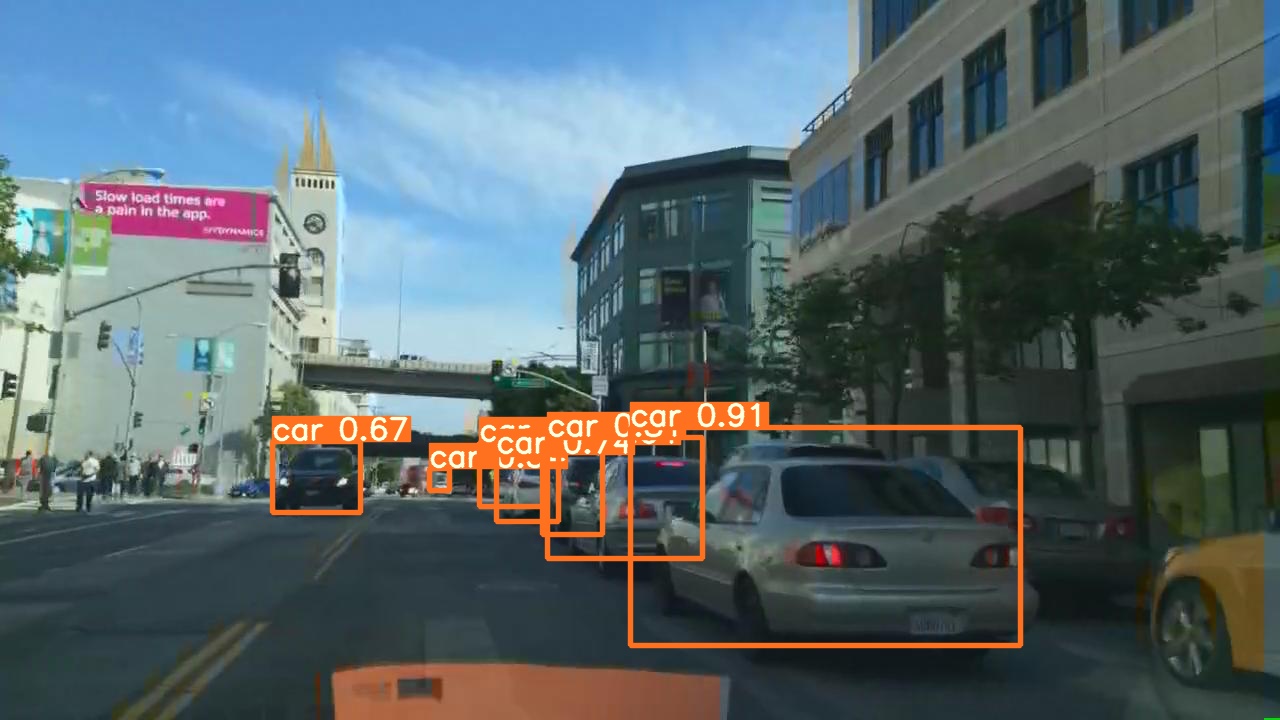}
    \caption{PiCANet}
\end{subfigure}
    \hfill
\begin{subfigure}{.3\linewidth}
    \centering
       \includegraphics[width=\linewidth]{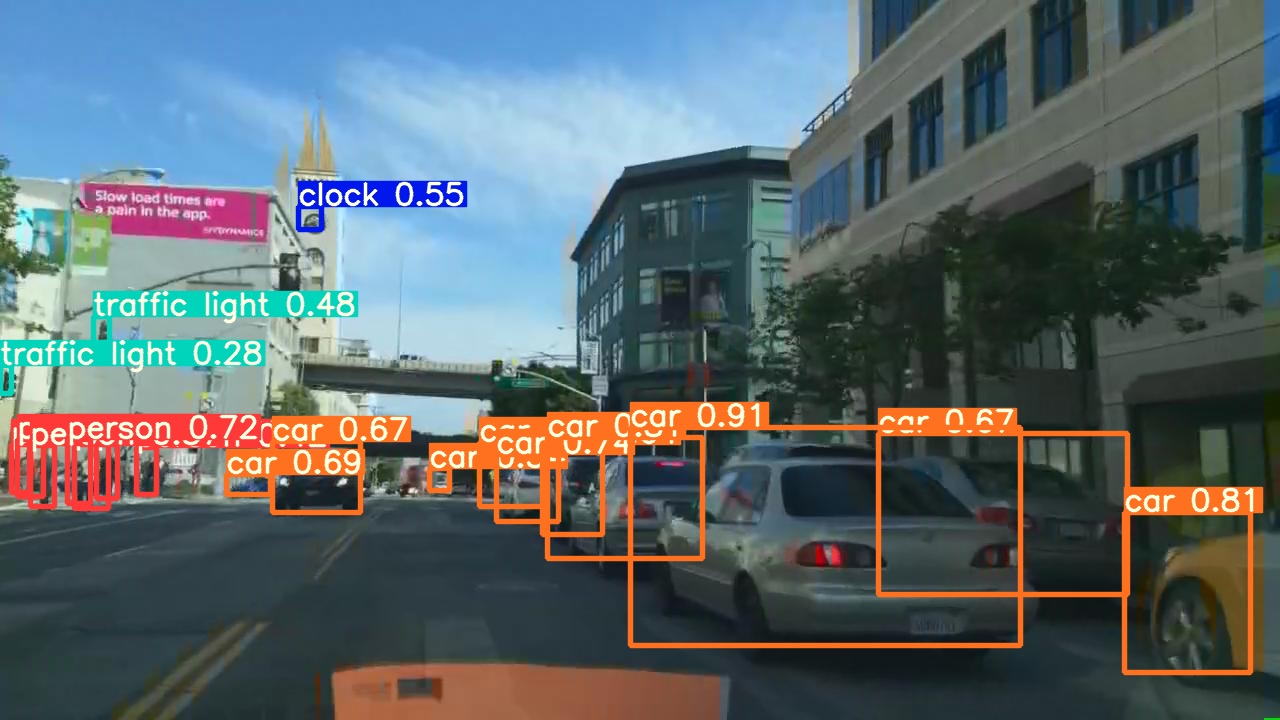}
    \caption{All objects}
\end{subfigure}
\caption{Comparison of attention-based object detection using different models. (a) Ground-truth attention; (b-d) Predictions using our framework with different backbones; (e-h) Predictions using models \cite{xia2018predicting,dreyeve2018,cornia2016deep,liu2018picanet}; (i) Object detection without driver attention. }
\label{fig:detected objects}
\end{figure}

\cref{fig:detected objects} shows one example of attention-based predicted objects using different models. The predicted objects
are framed with bounding boxes. The frame is taken from a video, where a vehicle drives towards a crossroad and passes waiting vehicles that are on the right lane of the road. Comparing (i) and (a), we see that the human driver pays attention to several objects but not most of the objects. Our models based on features from YOLOv5 as well as CenterTrack backbones predict all waiting vehicles as focused by drivers (in (b) and (d)), matching with the ground-truth (in (a)). BDD-A prediction focuses on a car on the oncoming lane and a church clock, missing a waiting car in the distance. Moreover, always predicting gaze at the vanishing point is a significant problem for driving saliency models. From this example, we can deduce that our model does not constantly predict the vanishing point in the street, whereas DR(eye)VE, ML-Net and PiCANet predict the object around the center point as critical. 

%From this, we also deduce that out model does not always predict the vanishing point in the street, because there is a vehicle behind the crossroad in the center, which was not predicted by our YOLOv5 model, even though it is predicted by YOLOv3, DR(eye)VE, ML-Net and PiCANet. Predicting gaze at the vanishing point is a significant problem for driving saliency models. The original object detection using YOLOv5 is shown in image (i).

\begin{figure}[]
    \centering
        \includegraphics[height=1in]{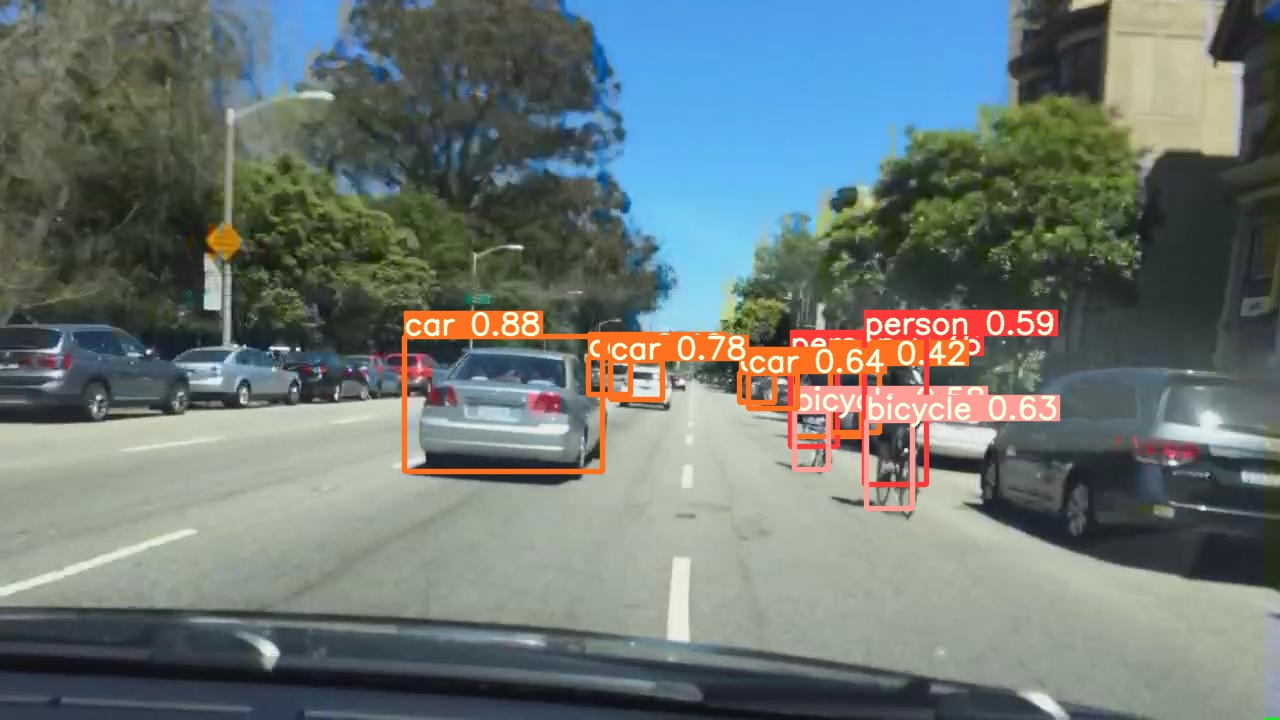}
        \includegraphics[height=1in]{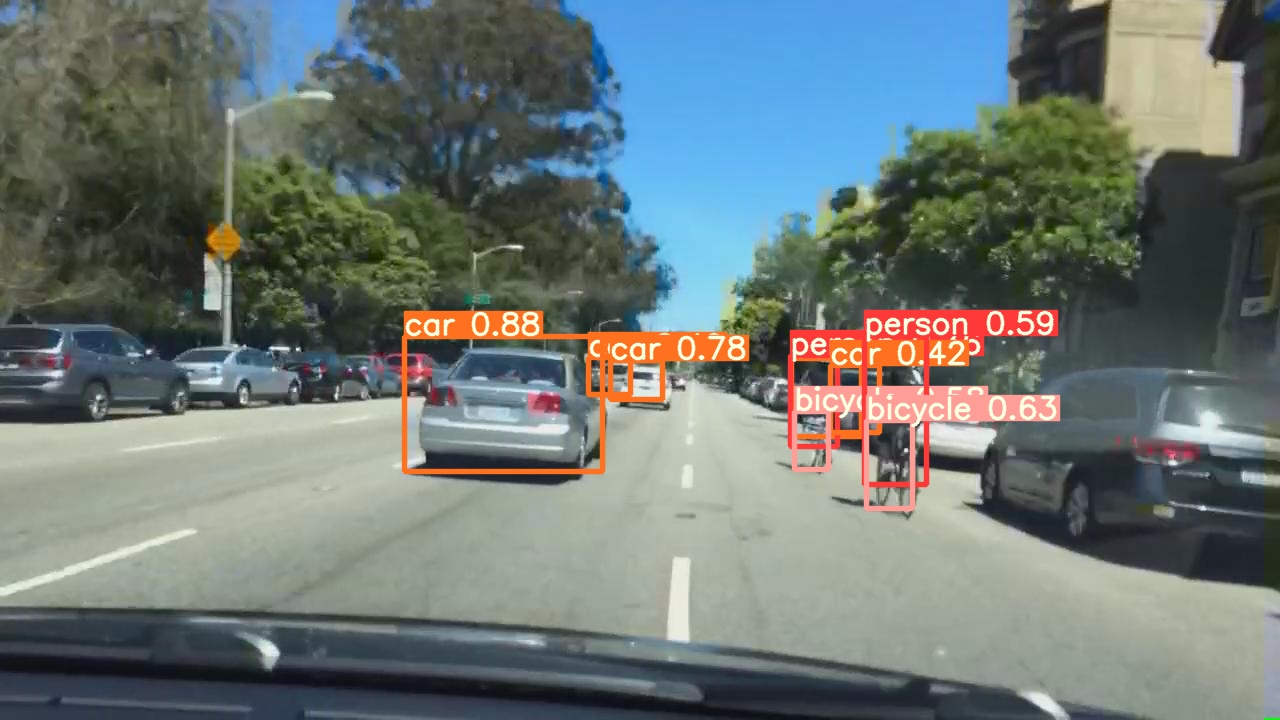}
        \includegraphics[height=1in]{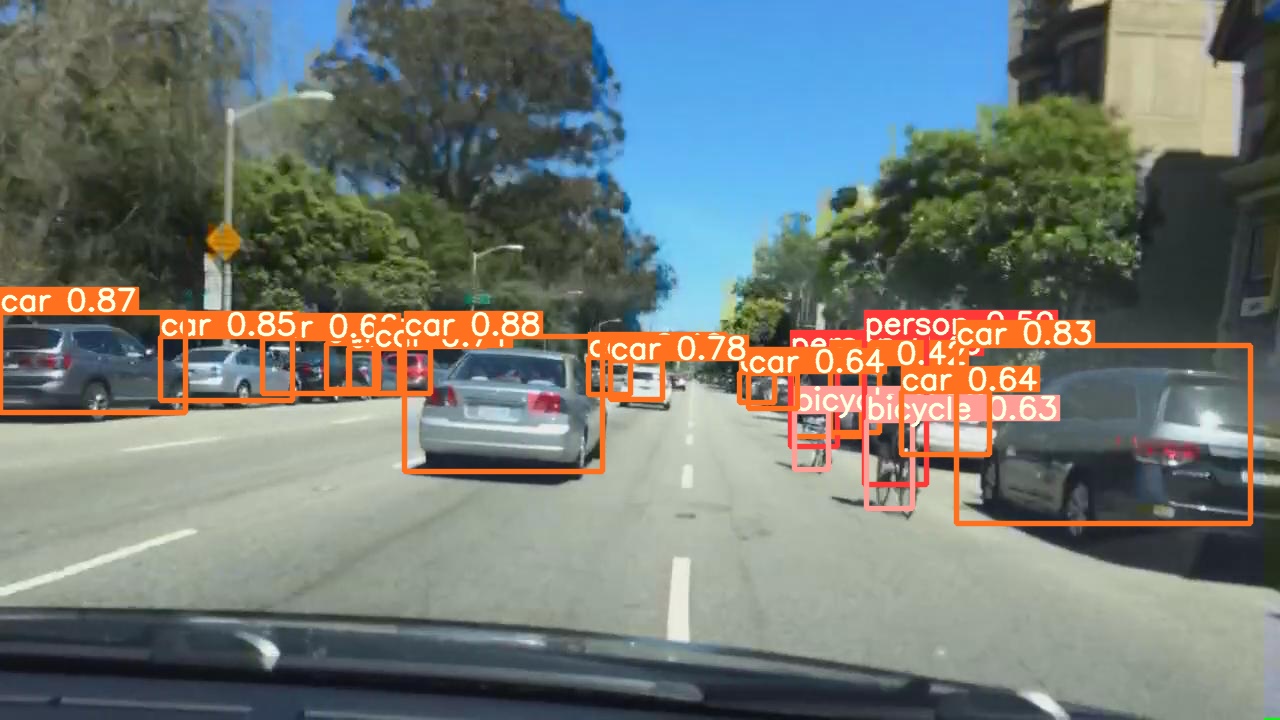}
        \includegraphics[height=1in]{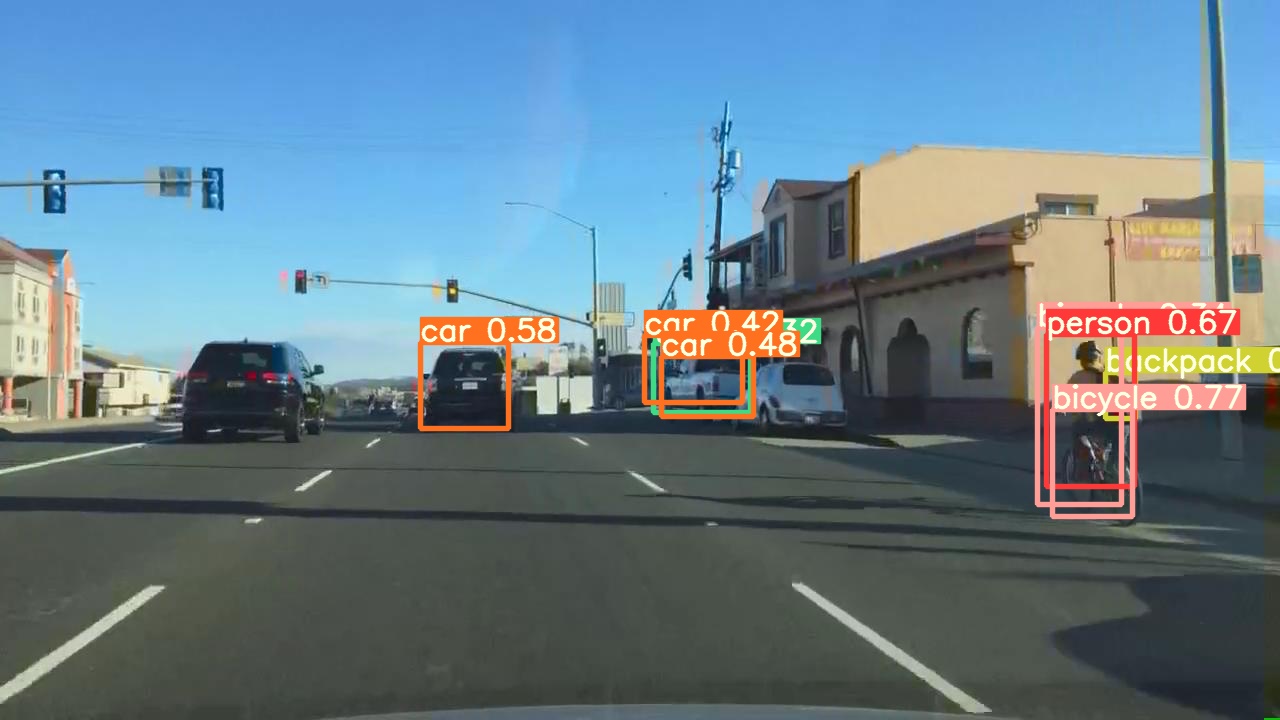}
        \includegraphics[height=1in]{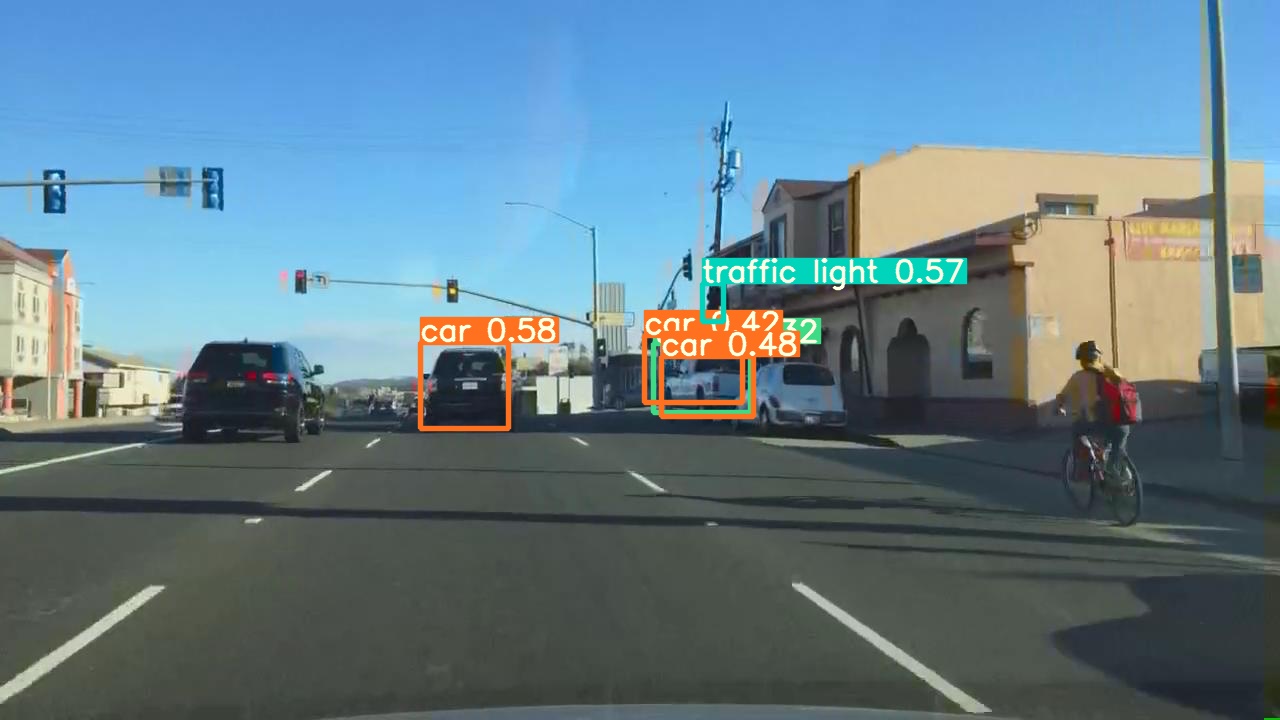}
        \includegraphics[height=1in]{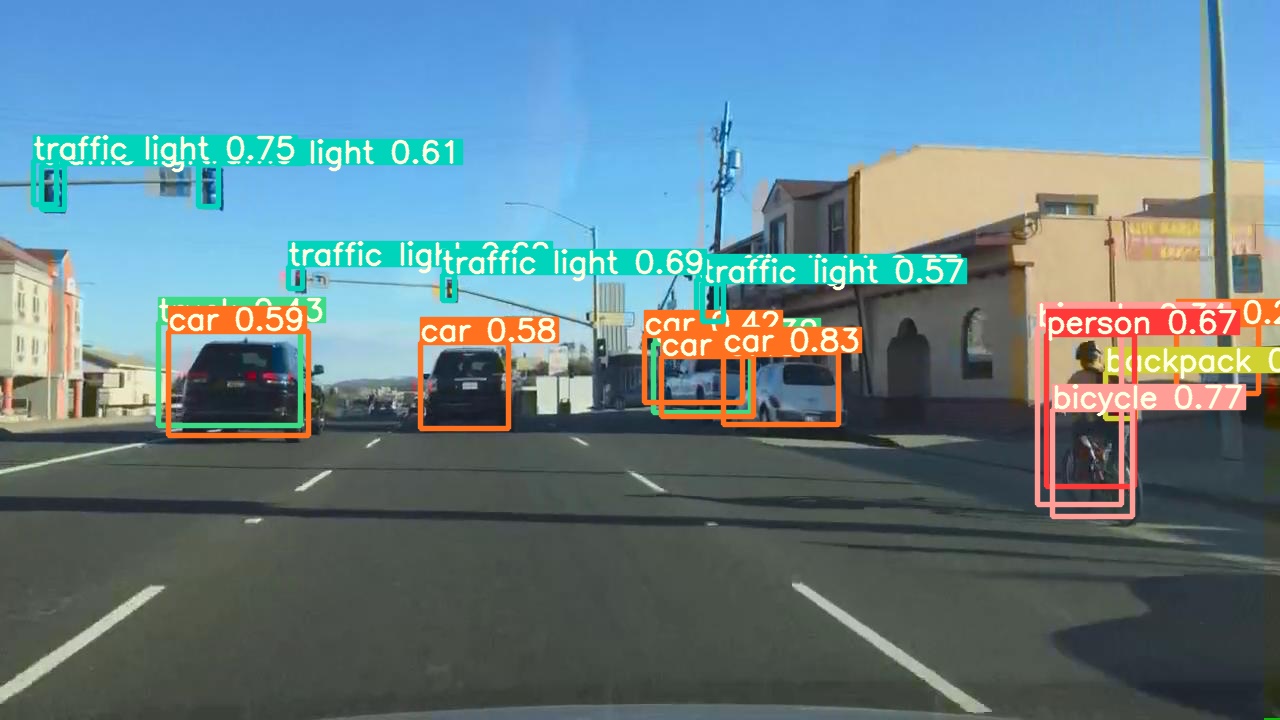}
    \caption{Comparison of our prediction, ground-truth in attention-based object detection and not using attention-based object detection on BDD-A test set. (Failed cases.) \textbf{Left}: Our prediction; \textbf{Middle}: Ground-truth; \textbf{Right}: Object detection without driver attention. Better view in colors.}
    \label{fig:failed cases}
\end{figure}

We also present two failed predictions of our YOLOv5 based model in \cref{fig:failed cases}. In the first row, the vehicle is changing lanes from the left to the middle to pass two cyclists. Our model correctly notices the cars in front of the vehicle as well as the cyclists. Directly in front of the cyclists, our model predicts wrongly parked cars to be critical compared to the ground-truth. Nevertheless, this is a good example for the effect of attention-based object detection. The vehicles in front and the cyclists, which might make it necessary to react, are detected, while the cars parked two lanes away are not detected. In the second row, a vehicle drives towards a crossroad with a traffic light turning red. Our model correctly predicts the vehicle braking in front on the same lane and a car parked on the right. But additionally, our model considers a cyclist on the right of the scene as critical. Although the cyclist is wrongly predicted, it shows that the predictions of our model are not limited to the center part of an image.

\subsection{Results on DR(eye)VE}
\subsubsection{Quantitative Results}\hfill \break
We tested our model on the DR(eye)VE dataset without further training to validate its generalization ability. We ran our YOLOv5 model in 16$\times$16 grids and compared it with DR(eye)VE, BDD-A, ML-Net and PiCANet. \naemi{As in the experiments on BDD-A, we computed the threshold individually with the ROC curves shown in \cref{sec:appendix6} and} evaluated the models on object-level with metrics $AUC$, precision, recall, $F_1$ and accuracy and on pixel-level with $D_{KL}$ and $CC$. The results are shown in \cref{tab:comparison in dreyeve}. The bottom-up models ML-Net and PiCANet achieved in our experimental setting better results than the top-down networks DR(eye)VE and BDD-A. Our model and PiCANet achieved the best results on object-level ($AUC = 0.88$) and outperformed all other models on pixel-level ($D_{KL} = 1.78$, $CC = 0.51$). Achieving good performance on DR(eye)VE shows that our model is not limited to the BDD-A dataset. 

\begin{table}[]
\caption{Comparison with other gaze models on DR(eye)VE dataset. On object-level, all models are evaluated with detected objects of YOLOv5. Our models uses 16$\times$16 grids. * indicates that the backbone is pretrained on COCO \cite{lin2014microsoft}, $\dagger$ on ImageNet \cite{deng2009imagenet} and $\ddagger$ on UCF101 \cite{soomro2012ucf101}.}
\label{tab:comparison in dreyeve}
\resizebox{.7\linewidth}{!}{% <------ Don't forget this %
\begin{tabular}{c|c|c|c|c|c|c|c|}
\cline{2-8}
& \multicolumn{5}{c|}{\textbf{Object-level}} 
& \multicolumn{2}{c|}{\textbf{Pixel-level}}  \\ \cline{2-8}
& \textit{AUC} &  \textit{Prec.  (\%)} & \textit{Recall  (\%)} & \textit{$F_1$  (\%)} & \textit{Acc  (\%)} & \textit{KL} & \textit{CC}   \\ \hline
\multicolumn{1}{|c|}{\textbf{Baseline}}  & 0.86   &  65.18  & 77.79  & 70.93  & 77.94 &   2.00 & 0.40    \\ \hline
\multicolumn{1}{|c|}{\textbf{BDD-A} \cite{xia2018predicting} $^\dagger$} & 0.84 &   71.63       &  73.34        & 72.48     &  78.38       &  2.07     &  0.46      \\ \hline
\multicolumn{1}{|c|}{\textbf{DR(eye)VE} \cite{dreyeve2018} $^\ddagger$} & 0.86 &    68.90        &    79.39        &  73.77   &    78.09    &  2.79 & 0.47    \\ \hline
\multicolumn{1}{|c|}{\textbf{ML-Net} \cite{cornia2016deep}$^\dagger$}  &  0.87  &  69.74 & 79.73  & 74.40 & 78.71  &  2.17  &   0.45  \\ \hline
\multicolumn{1}{|c|}{\textbf{PiCANet} \cite{liu2018picanet}$^\dagger$} & 0.88 & 73.90         &  81.48 & 77.50  &   81.64    & 2.36  & 0.41 \\ \hline
%\multicolumn{1}{|c|}{\textbf{Ours (CenterTrack)}}  &  &  &   &    &  &  &       \\ \hline
%\multicolumn{1}{|c|}{\textbf{Ours (YOLOv3)}} &   &   & &   &    &  &    \\ \hline
\multicolumn{1}{|c|}{\textbf{Ours (YOLOv5)*}} & 0.88 &75.33 & 78.73   & 76.99  & 81.74 & 1.78 &    0.51    \\ \hline
\end{tabular}
}
\end{table}

\subsubsection{Qualitative Results}
\begin{figure*}[t]
    \centering
        \includegraphics[width = 0.3\textwidth, height=1.1in]{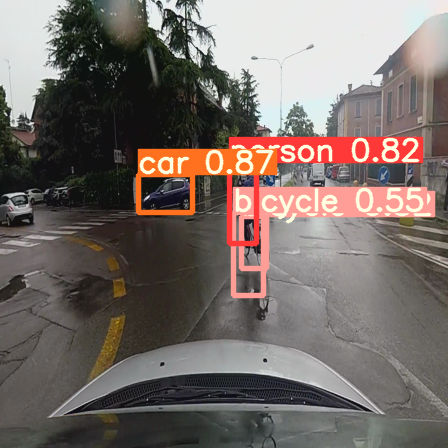}
        \includegraphics[width = 0.3\textwidth, height=1.1in]{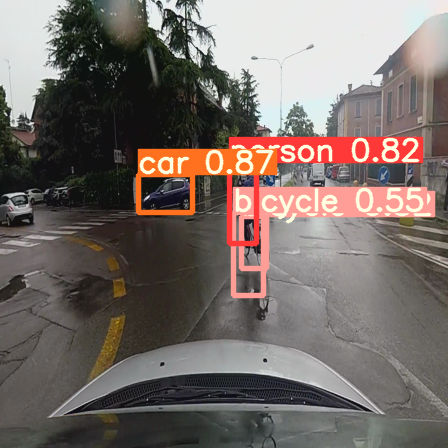}
        \includegraphics[width = 0.3\textwidth, height=1.1in]{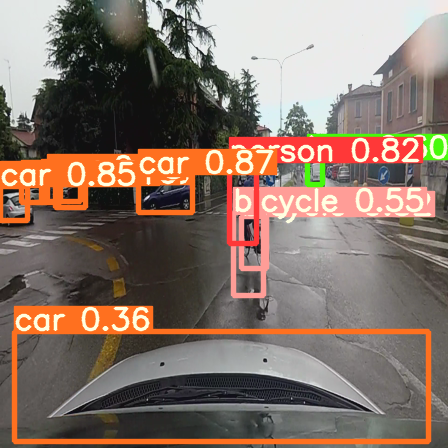}
        \includegraphics[width = 0.3\textwidth, height=1.1in]{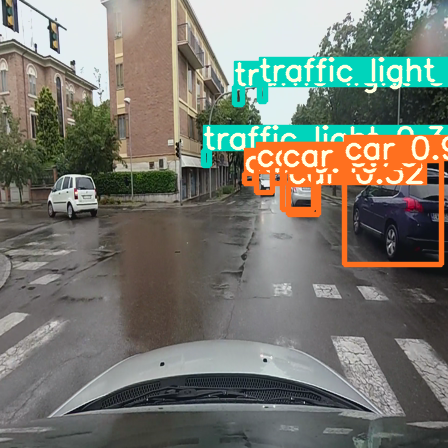}
        \includegraphics[width = 0.3\textwidth, height=1.1in]{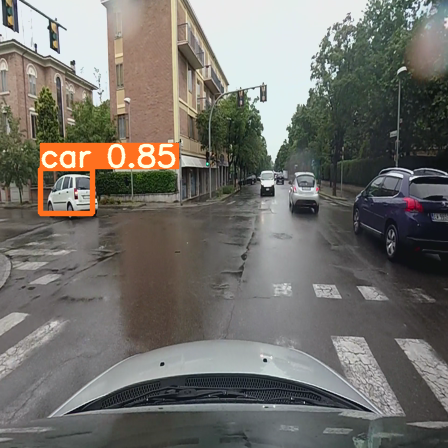}
        \includegraphics[width = 0.3\textwidth, height=1.1in]{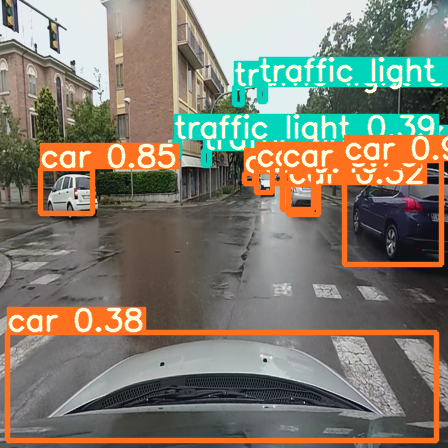}
    \caption{Comparison of our prediction, ground-truth in attention-based object detection and not using attention-based object detection on the DR(eye)VE testset ($Th = 0.4$ to better illustrate the wrongly predicted attention region in the failed case). (Second line is failed case.) \textbf{Left}: Our prediction; \textbf{Middle}: Ground-truth; \textbf{Right}: Object detection without driver attention. Better view in colors.}
    \label{fig:dreyeve examples}
\end{figure*}

\hfill \break
\cref{fig:dreyeve examples} shows two examples of our attention-based object prediction model on the DR(eye)VE dataset. The frames in the first row belong to a video sequence where the driver follows the road in a left curve. Our model (left) detects the cyclist driving in front of the car and a vehicle waiting on the right to merge. Other cars further away were not predicted as focused, thus it matches the ground-truth (middle). In the second row, we can see a frame where the driver wants to turn left. Our model (left) predicts the cars and traffic lights on the road straight ahead, whereas the ground-truth (middle) covers a car turning left. This example underlines the difficulty of predicting drivers' attention when it depends on different driving goals \yao{\cite{yarbus1967eye}}. 

\section{Discussion}
In this section, we first show our LSTM-variant architecture and discuss the results to address the challenges of using temporal information in this task. Then, we deliberate other limitations of the current project.

\subsection{\yao{Modelling with LSTM-Layer}}
To extend our framework into a video-based prediction, we added one LSTM-layer (Long Short-Term Memory \cite{hochreiter1997long}) with 256 as the size of the hidden state before the dense layer in the gaze prediction network. The input for this network is an eight-frame video clip. %The last output of the LSTM layer is the region vector of dimension $m \times n$, where $m$ and $n$ are the grid size.
We tested our extended architecture using the same configuration described in the last section (i.e., 16$\times$16 grids with $Th$ of 0.5) and achieved the following results on the BDD-A dataset:
    \\\indent \textbf{Object Detection}: $AUC$ = 0.85, Precision = 73.13\%,  Recall = 70.44\%,  \\ 
    \indent \indent \indent \indent \indent \indent
    \indent \indent \indent 
    %%% looks dirty but works...
    $F_1$ score = 71.76\%, Accuracy = 77.83\%
    \\\indent \textbf{Saliency Prediction}: $D_{KL}$ = 1.17, $CC$ = 0.60

The above results are similar to our model without the LSTM-layer, both achieved $AUC = 0.85$ and $CC = 0.60$. It is worth mentioning that the sequence length (from 2 to 16) had no significant influence on the performance. (See \cref{sec:appendix4} for more results.) Similarly, \cite{xia2018predicting} also observes that using LSTM-layers cannot improve the performance in driver gaze prediction but rather introduces center biases into the prediction.
In summary, more frames do not increase the information gain. %One possible reason behind this is that the visual image features of each frame inside a sequence do not vary much, since the environment on the street does not change very quickly in the video clips from the current datasets. 
One possible reason behind this bias is that using an LSTM-layer ignores the spatial information, since the extracted features given to the LSTM-layer are reshaped to vectors. 
% Increasing input sequence length can probably introduce more information, especially temporal changes between different frames, but it raises the computation costs as well. 
Therefore, in the context of our future work, we would like to analyze the integration of other modules that include temporal information, such as the convolutional LSTM (convLSTM) \cite{xingjian2015convolutional}. Using convLSTM can capture the temporal information of each spatial region and predict the new features for it based on the past motion information inside the region. 
For example, \cite{xia2018predicting,rong2020driver} validate that convLSTM helps capture the spatial-temporal information for driver attention/action predictions. Another proposal is to use 3D CNN to get the spatial-temporal features. For instance, \cite{dreyeve2018} deploys 3D convolutional layers that takes a sequence of frames as input in predicting the driver's attention.

\subsection{\yao{Limitations and Future Work}}
One limitation of current projects is that all current models have a central bias in their prediction.
\yao{This effect stems from the ground-truth data because human drivers naturally look at the center part of the street, creating very unbalanced data: 74.2\% of all focused objects on BDD-A come from the central bias area as shown in the baseline in \Cref{fig:saliency prediction}. The central bias reflects natural human behavior and is even enhanced in the saliency models proposed by K\"ummerer et al.~\cite{kummerer2014deep, kummerer2016deepgaze}. Although our model predicts objects in the margin area of the scene as shown in our qualitative examples, the center is often prioritized.} Our model has an $F_1$ score of 81.7\% inside of the center area, while it only reaches 34.8\% in $F_1$ outside of the center area. PiCANet, which achieves the best result among all models, has better $F_1$ scores outside (44.0\%) and inside of the center (82.7\%), however, its performance inside of the center is dominant. We intend to improve the model prediction outside of the center but still keep the good performance in the center area in the future. In the context of autonomous driving, it would be also essential to test the generalization ability on other datasets, which are not limited to just the gaze map data. Since drivers also rely on peripheral vision, they do not focus on every relevant object around them. Using other datasets that additionally highlight objects based on semantic information (e.g., \cite{pal2020looking}) could increase the applicability for finding task-relevant objects.

\yao{All models in the experiments are trained on saliency maps derived from driver gaze. These salient features are related to regions of interest where a task-relevant
object should be located, thus reflecting top-down features \cite{oliva2003top}. However, these features are currently extracted from the visual information given by camera images. The context of driving tasks can still be enhanced by adding more input information, since human top-down feature selection mechanisms require comprehensive understanding of the task that is outside the realm of visual perception. Concretely, the driver’s attention can be affected by extrinsic factors such as road conditions, or intrinsic factors such as driver intentions based on driving destinations. These factors, along with traffic information, form the driver attention as well as gaze patterns. Unfortunately, the current dataset used for our model training does not provide this additional input. For the future work, we will consider incorporating GPS and Lidar sensor information, which can provide more insights of tasks to better predict driver attention.}

%The results show that there is no advantage to using a video clip. The lack of consistency in critical region detection leads to lower recall and a lower F1 score. If one region is not detected as critical at the beginning of a video clip, this region will not be detected in the end. This leads to lower recall.

%\textcolor{blue}{other limitations: center bias, etc.}

\section{Conclusion}
In this paper, we propose a novel framework to detect human attention-based objects in driving scenarios. Our framework predicts driver attention saliency maps and detects objects inside the predicted area. This detection is achieved by using the same backbone (feature encoder) for both tasks, and the saliency map is predicted in grids. In doing so, our framework is highly computation-efficient. Comprehensive experiments on two driver attention datasets, BDD-A and DR(eye)VE, show that our framework achieves competitive results in the saliency map prediction and object detection compared to other state-of-the-art models while reducing computational costs.
% Conclusion and Further work: yolox as object detection network; maybe include more information (like semantic-gaze) ; avoid center bias; use temporal information  

\section{Acknowledgments}
We acknowledge the support by Cluster of Excellence - Machine Learning: New Perspectives for Science, EXC number 2064/1 - Project number 390727645.

% For ETRA Articles 137+, use
\received{November 2021}
\received[revised]{January 2022}
\received[accepted]{April 2022}

\bibliographystyle{ACM-Reference-Format}
\bibliography{sample-base}


\appendix
\section{Visualization of the ROC curves}
\label{sec:appendix6}
\begin{figure}[h]
    \centering
        \includegraphics[width = .4\textwidth]{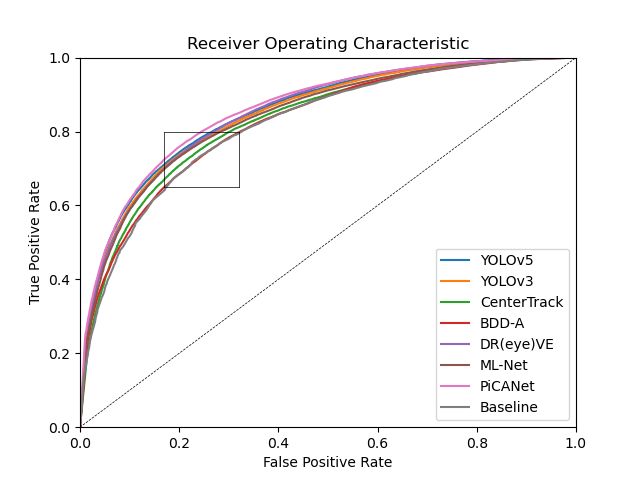}
         \includegraphics[width = .4\textwidth]{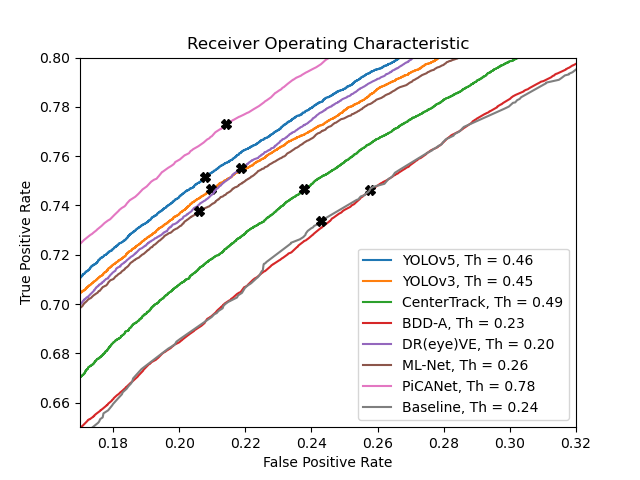}
    \caption[ROC curves and computed thresholds on the BDD-A test set.]{ROC curves and computed thresholds on the BDD-A test set. On the right, the curves are zoomed in and the 
    points that belong to the computed thresholds are marked.}
    \label{fig:roc BDDA}
\end{figure}

\begin{figure}[h]
    \centering
        \includegraphics[width = .4\textwidth]{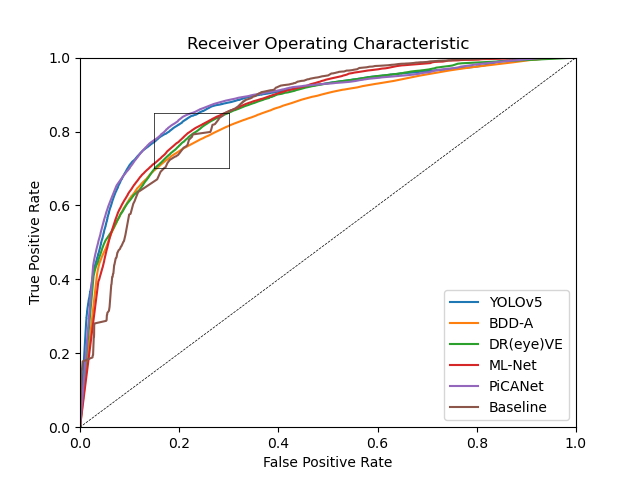}
         \includegraphics[width = .4\textwidth]{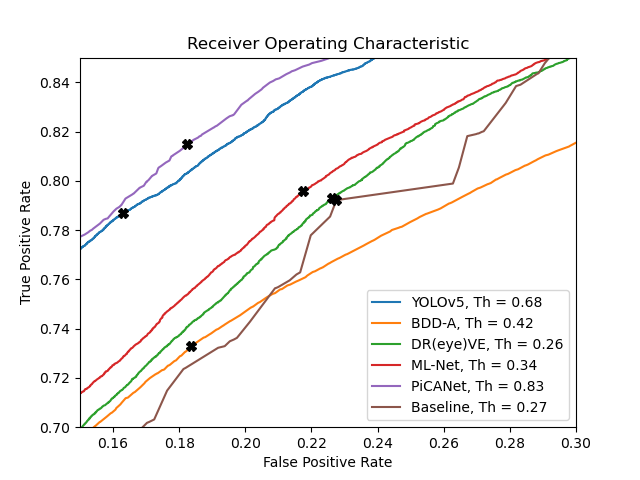}

    \caption{ROC curves and computed thresholds on the DR(eye)VE test set. On the right, the curves are zoomed in and the 
    points that belong to the computed thresholds are marked.}
    \label{fig:roc DREYEVE}
\end{figure}

In \cref{fig:roc BDDA} and \cref{fig:roc DREYEVE} we show the ROC curves and computed thresholds for all models on the BDD-A and DR(eye)VE test sets.

\section{More Quantitative Results}
\subsection{Results of Other Thresholds on BDD-A}
\label{sec:appendix1}
Our models always achieve high $F_1$ scores in different $Th$, indicating that our models have relatively good performance in precision and recall scores at the same time. PiCANet is more unbalanced in recall and precision compared to other models. The accuracy scores are influenced by the $Th$ values, however, the highest accuracy 78.55\% is achieved by our YOLOv5-based model when $Th$ is set to 0.6. \\

%\noindent
%\vfill
\begin{minipage}[c]{0.45\linewidth}
\centering
%\begin{table}[h!]
\captionof{table}{Comparison of different models with Th = 0.3 on BDD-A dataset. Results are shown in \% and for all metrics, a higher value indicates better performance.}
%\caption{Comparison of different Th for YOLOv5 grid 16$\times$16 models on BDD-A test set. For all metrics the higher value represents better performance.}
\label{tab:different Th 03}
\resizebox{\linewidth}{!}{% <------ Don't forget this %
\begin{tabular}{c|c|c|c|c|}
\cline{2-5}
& \textbf{Prec} & \textbf{Recall} & $\mathbf{F_1}$ & \textbf{Acc }       \\ \hline
\multicolumn{1}{|c|}{\textbf{BDD-A}}   &  68.88  &  69.43     &   69.16  &    75.24       \\ \hline
\multicolumn{1}{|c|}{\textbf{DR(eye)VE}}   & 75.32   &  66.42     & 70.59    &     77.87      \\ \hline

\multicolumn{1}{|c|}{\textbf{ML-Net}}   & 72.84   &   70.43    &    71.61 &    77.68       \\ \hline

\multicolumn{1}{|c|}{\textbf{PiCANet}}   &  43.61  &   99.36    &  60.61   &    48.36       \\ \hline

\multicolumn{1}{|c|}{\textbf{Ours (CenterTrack)}}   &  61.19  &    83.11   & 70.49    &    72.17       \\ \hline

\multicolumn{1}{|c|}{\textbf{Ours (YOLOv3) }}   & 63.97   &  82.15     &  71.93   &    74.36       \\ \hline

\multicolumn{1}{|c|}{\textbf{Ours (YOLOv5)}} &   63.76          &    83.33      &  72.24    &   74.39      \\ \hline
\end{tabular}
}
\end{minipage} \hfill \begin{minipage}[c]{0.45\linewidth}
\centering
%\begin{table}[h!]
\captionof{table}{Comparison of different models with Th = 0.4 on BDD-A dataset. Results are shown in \% and for all metrics, a higher value indicates better performance.}
%\caption{Comparison of different Th for YOLOv5 grid 16$\times$16 models on BDD-A test set. For all metrics the higher value represents better performance.}
\label{tab:different Th 04}
\resizebox{\linewidth}{!}{% <------ Don't forget this %
\begin{tabular}{c|c|c|c|c|}
\cline{2-5}
& \textbf{Prec} & \textbf{Recall} & $\mathbf{F_1}$ & \textbf{Acc }       \\ \hline
\multicolumn{1}{|c|}{\textbf{BDD-A}}   &  72.68  &   63.44    &  67.75   &    75.85       \\ \hline
\multicolumn{1}{|c|}{\textbf{DR(eye)VE}}   &  78.99  &   59.95    &  68.16   &    77.61       \\ \hline

\multicolumn{1}{|c|}{\textbf{ML-Net}}   &  77.50  &  62.79     &    69.37 &      77.83     \\ \hline

\multicolumn{1}{|c|}{\textbf{PiCANet}}   & 47.15   &  98.26     & 63.72   &     55.27      \\ \hline

\multicolumn{1}{|c|}{\textbf{Ours (CenterTrack)}}   &  65.52  &    77.86   &   71.15  &     74.76      \\ \hline

\multicolumn{1}{|c|}{\textbf{Ours (YOLOv3) }}   & 68.16   &  77.02     &  72.32   &  76.43         \\ \hline

\multicolumn{1}{|c|}{\textbf{Ours (YOLOv5)}}  &  68.11          &     78.36       &  72.88    &   76.68      \\ \hline
\end{tabular}
}
\end{minipage}

\ \\

%\vfill
\begin{minipage}[c]{0.45\textwidth}
\centering
%\begin{table}[h!]
\captionof{table}{Comparison of different models on BDD-A dataset with Th = 0.5. Results are shown in \% and for all metrics, a higher value indicates better performance.}
%\caption{Comparison of different Th for YOLOv5 grid 16$\times$16 models on BDD-A test set. For all metrics the higher value represents better performance.}
\label{tab:different Th 05}
%\resizebox{\linewidth}{!}{% <------ Don't forget this %
\resizebox{!}{.2\linewidth}{
\begin{tabular}{c|c|c|c|c|}
\cline{2-5}
& \textbf{Prec} & \textbf{Recall} & $\mathbf{F_1}$ & \textbf{Acc }       \\ \hline
\multicolumn{1}{|c|}{\textbf{BDD-A}}   & 75.84  & 57.44     &  65.37 &     75.67      \\ \hline
\multicolumn{1}{|c|}{\textbf{DR(eye)VE}}   &  81.84  &  54.38    &   65.35  &      76.94     \\ \hline

\multicolumn{1}{|c|}{\textbf{ML-Net}}   &  80.96  &    55.75  &  66.03 &    77.07       \\ \hline

\multicolumn{1}{|c|}{\textbf{PiCANet}}   &   51.26 &     95.98  &  66.83   &  61.90       \\ \hline

\multicolumn{1}{|c|}{\textbf{Ours (CenterTrack)}}   & 69.29   &  72.19     &  70.71   &     76.09      \\ \hline

\multicolumn{1}{|c|}{\textbf{Ours (YOLOv3) }}   & 72.14  &  72.23    &  72.18   &    77.74       \\ \hline

\multicolumn{1}{|c|}{\textbf{Ours (YOLOv5)}} &  71.98        &    73.31       &  72.64    &   77.92      \\ \hline
\end{tabular}
}
\end{minipage} \hfill \begin{minipage}[c]{0.45\linewidth}
\centering
%\begin{table}[h!]
\captionof{table}{Comparison of different models on BDD-A dataset with Th = 0.6. Results are shown in \% and for all metrics, a higher value indicates better performance.}
%\caption{Comparison of different Th for YOLOv5 grid 16$\times$16 models on BDD-A test set. For all metrics the higher value represents better performance.}
\label{tab:different Th 06}
\resizebox{\linewidth}{!}{% <------ Don't forget this 
\begin{tabular}{c|c|c|c|c|}
\cline{2-5}
& \textbf{Prec} & \textbf{Recall} & $\mathbf{F_1}$ & \textbf{Acc }       \\ \hline
\multicolumn{1}{|c|}{\textbf{BDD-A}}   & 78.84   & 51.41      &   62.23  &     75.05      \\ \hline
\multicolumn{1}{|c|}{\textbf{DR(eye)VE}}   &  84.13  &   49.57    &   62.39  &      76.10     \\ \hline

\multicolumn{1}{|c|}{\textbf{ML-Net}}   &  83.53  &    48.80   &   61.61  &    75.68       \\ \hline

\multicolumn{1}{|c|}{\textbf{PiCANet}}   &   56.31 &     92.30  &  69.95   &   68.29        \\ \hline

\multicolumn{1}{|c|}{\textbf{Ours (CenterTrack)}}   & 73.13   &  66.12     &  69.45   &     76.74      \\ \hline

\multicolumn{1}{|c|}{\textbf{Ours (YOLOv3) }}   & 75.71   &  66.68     &  70.91   &    78.12       \\ \hline

\multicolumn{1}{|c|}{\textbf{Ours (YOLOv5)}} &  75.81          &     68.09       &  71.74    &   78.55       \\ \hline
\end{tabular}
}
\end{minipage} 
\\

\begin{minipage}[c]{0.45\linewidth}
\centering
\captionof{table}{Comparison of different models on BDD-A dataset with Th = 0.7. Results are shown in \% and for all metrics, a higher value indicates better performance.}
%\caption{Comparison of different Th for YOLOv5 grid 16$\times$16 models on BDD-A test set. For all metrics the higher value represents better performance.}
\label{tab:different Th 07}
\resizebox{\linewidth}{!}{% <------ Don't forget this %
\begin{tabular}{c|c|c|c|c|}
\cline{2-5}
& \textbf{Prec} & \textbf{Recall} & $\mathbf{F_1}$ & \textbf{Acc }       \\ \hline
\multicolumn{1}{|c|}{\textbf{BDD-A}}   &  81.90  &   45.57    &  58.56   &     74.21      \\ \hline
\multicolumn{1}{|c|}{\textbf{DR(eye)VE}}   & 86.14   &    44.34   &   58.55  &    74.89       \\ \hline

\multicolumn{1}{|c|}{\textbf{ML-Net}}   & 85.85   &   42.31    &   56.69  &    74.15       \\ \hline

\multicolumn{1}{|c|}{\textbf{PiCANet}}   &  63.10  & 85.88      &  72.75   &       74.28    \\ \hline

\multicolumn{1}{|c|}{\textbf{Ours (CenterTrack)}}   &  76.91  &     59.52  &  67.11   &   76.67        \\ \hline

\multicolumn{1}{|c|}{\textbf{Ours (YOLOv3) }}   & 79.39   &   60.44    &  68.63   &     77.91      \\ \hline

\multicolumn{1}{|c|}{\textbf{Ours (YOLOv5)}} & 79.61        &     62.04       &  69.73    &   78.47       \\ \hline
\end{tabular}
}
\end{minipage}

\subsection{Results of Other Thresholds on DR(eye)VE}
\label{sec:appendix2}
Our model achieves the best $F_1$ socre of 76.94\% and accuracy of 81.9\%, while the best $F_1$ score and accuracy scores among other models are 74.24\% and 79.68\% respectively, which validates the good performance of our model in the attention-based object detection task.

\begin{minipage}[c]{0.45\linewidth}
\centering
%\begin{table}[h!]
\captionof{table}{Comparison of different models on DR(eye)VE dataset with Th = 0.3. Results are shown in \% and for all metrics, a higher value indicates better performance.}
%\caption{Comparison of different Th for YOLOv5 grid 16$\times$16 models on BDD-A test set. For all metrics the higher value represents better performance.}
\label{tab:different Th Dreyeve 03}
\resizebox{\linewidth}{!}{% <------ Don't forget this %
\begin{tabular}{c|c|c|c|c|}
\cline{2-5}
& \textbf{Prec} & \textbf{Recall} & $\mathbf{F_1}$ & \textbf{Acc }       \\ \hline
\multicolumn{1}{|c|}{\textbf{BDD-A}}   & 65.94   &  78.83     &  71.81   &     75.98      \\ \hline
\multicolumn{1}{|c|}{\textbf{DR(eye)VE}}   &  70.34 &  76.95    &  73.50   &   78.46      \\ \hline

\multicolumn{1}{|c|}{\textbf{ML-Net}}   &  67.98  &   81.77    &   74.24  &    77.98       \\ \hline

\multicolumn{1}{|c|}{\textbf{PiCANet}}   &  42.34  &   98.98    &   59.31  &    47.31       \\ \hline

\multicolumn{1}{|c|}{\textbf{Ours (YOLOv5)}}   &  58.08  &  91.25     &   70.98  &   71.04       \\ \hline
\end{tabular}
}
\end{minipage} \hfill \begin{minipage}[c]{0.45\linewidth}
\centering
%\begin{table}[h!]
\captionof{table}{Comparison of different models on DR(eye)VE dataset with Th = 0.4. Results are shown in \% and for all metrics, a higher value indicates better performance.}
%\caption{Comparison of different Th for YOLOv5 grid 16$\times$16 models on BDD-A test set. For all metrics the higher value represents better performance.}
\label{tab:different Th Dreyeve 04}
\resizebox{\linewidth}{!}{% <------ Don't forget this %
\begin{tabular}{c|c|c|c|c|}
\cline{2-5}
& \textbf{Prec} & \textbf{Recall} & $\mathbf{F_1}$ & \textbf{Acc }       \\ \hline
\multicolumn{1}{|c|}{\textbf{BDD-A}}   & 70.82   &  74.16     &   72.45  &   78.12        \\ \hline
\multicolumn{1}{|c|}{\textbf{DR(eye)VE}}   &  73.57  &   71.54    &   72.54  &    78.98       \\ \hline

\multicolumn{1}{|c|}{\textbf{ML-Net}}   & 71.85   &   76.23    &   73.97  &   79.19        \\ \hline

\multicolumn{1}{|c|}{\textbf{PiCANet}}   &  46.83  &  95.81     &  62.91   &    56.16       \\ \hline

\multicolumn{1}{|c|}{\textbf{Ours (YOLOv5)}}   &  62.81  &  89.19     &   73.71  &    75.31       \\ \hline
\end{tabular}
}
\end{minipage}

\ \\

\begin{minipage}[c]{0.45\linewidth}
\centering
%\begin{table}[h!]
\captionof{table}{\naemi{(ADDED)} Comparison of different models on DR(eye)VE dataset with Th = 0.5. Results are shown in \% and for all metrics, a higher value indicates better performance.}
%\caption{Comparison of different Th for YOLOv5 grid 16$\times$16 models on BDD-A test set. For all metrics the higher value represents better performance.}
\label{tab:different Th Dreyeve 05}
\resizebox{\linewidth}{!}{% <------ Don't forget this %
\begin{tabular}{c|c|c|c|c|}
\cline{2-5}
& \textbf{Prec} & \textbf{Recall} & $\mathbf{F_1}$ & \textbf{Acc }       \\ \hline
\multicolumn{1}{|c|}{\textbf{BDD-A}}   & 74.58   &  69.53    &  71.97   &    78.98      \\ \hline
\multicolumn{1}{|c|}{\textbf{DR(eye)VE}}   & 76.21   &  66.46   &   71.01  &    78.94       \\ \hline

\multicolumn{1}{|c|}{\textbf{ML-Net}}   & 75.48   &   71.02    &   73.19 &    79.80      \\ \hline

\multicolumn{1}{|c|}{\textbf{PiCANet}}   & 51.30   & 93.92    &  66.36   &   63.05      \\ \hline

\multicolumn{1}{|c|}{\textbf{Ours (YOLOv5)}}   &  68.33  &  85.83    &  76.08  &   79.06      \\ \hline
\end{tabular}
}
\end{minipage} \hfill\begin{minipage}[c]{0.45\linewidth}
\centering
%\begin{table}[h!]
\captionof{table}{Comparison of different models on DR(eye)VE dataset with Th = 0.6. Results are shown in \% and for all metrics, a higher value indicates better performance.}
%\caption{Comparison of different Th for YOLOv5 grid 16$\times$16 models on BDD-A test set. For all metrics the higher value represents better performance.}
\label{tab:different Th Dreyeve 06}
\resizebox{\linewidth}{!}{% <------ Don't forget this %
\begin{tabular}{c|c|c|c|c|}
\cline{2-5}
& \textbf{Prec} & \textbf{Recall} & $\mathbf{F_1}$ & \textbf{Acc }       \\ \hline
\multicolumn{1}{|c|}{\textbf{BDD-A}}   & 77.34   &  65.54     &  70.95   &    79.17       \\ \hline
\multicolumn{1}{|c|}{\textbf{DR(eye)VE}}   & 79.25   &   61.67    &   69.37  &    78.86       \\ \hline

\multicolumn{1}{|c|}{\textbf{ML-Net}}   & 78.43   &   66.84    &   72.17  &    80.00       \\ \hline

\multicolumn{1}{|c|}{\textbf{PiCANet}}   & 56.95   & 92.19      &  70.40   &    69.92       \\ \hline

\multicolumn{1}{|c|}{\textbf{Ours (YOLOv5)}}   &  71.90  &  82.26     &  76.73   &    80.64       \\ \hline
\end{tabular}
}
\end{minipage} %\hfill 

\ \\

\begin{minipage}[c]{0.45\linewidth}
\centering
%\begin{table}[h!]
\captionof{table}{Comparison of different models on DR(eye)VE dataset with Th = 0.7. Results are shown in \% and for all metrics, a higher value indicates better performance.}
%\caption{Comparison of different Th for YOLOv5 grid 16$\times$16 models on BDD-A test set. For all metrics the higher value represents better performance.}
\label{tab:different Th Dreyeve 07}
\resizebox{\linewidth}{!}{% <------ Don't forget this %
\begin{tabular}{c|c|c|c|c|}
\cline{2-5}
& \textbf{Prec} & \textbf{Recall} & $\mathbf{F_1}$ & \textbf{Acc }       \\ \hline
\multicolumn{1}{|c|}{\textbf{BDD-A}}   & 79.74   &   61.61    &  69.51   &    79.03       \\ \hline
\multicolumn{1}{|c|}{\textbf{DR(eye)VE}}   &  81.88  &  57.21     &  67.35   &    78.48       \\ \hline

\multicolumn{1}{|c|}{\textbf{ML-Net}}   & 80.70   &  62.61     &   70.51  &    79.68       \\ \hline

\multicolumn{1}{|c|}{\textbf{PiCANet}}   &  62.88  &  89.49    &  73.86   &    75.42       \\ \hline

\multicolumn{1}{|c|}{\textbf{Ours (YOLOv5)}}   & 76.09   &   77.80    &  76.94   &   81.90        \\ \hline

\end{tabular}
}
\end{minipage}

\subsection{Results of Our YOLOv3- and CenterTrack-based Models} 
\label{sec:appendix3}
For a fair comparison, we computed object-level metrics with the detected objects of YOLOv5 for all models in \cref{sec:results}. In \cref{tab:other bb}, we show the object-level results for our $16\times16$ grids YOLOv3 and CenterTrack based models using their detected objects. 

\begin{table}[h!]
\centering
%\begin{table}[h!]
\captionof{table}{Comparison of different models on BDD-A dataset with own detected objects (Th = 0.5). For all metrics a higher value indicates better performance.}

\label{tab:other bb}
\resizebox{.6\linewidth}{!}{% <------ Don't forget this %
\begin{tabular}{c|c|c|c|c|c|}
\cline{2-6}
& \textbf{AUC} & \textbf{Prec (\%)} & \textbf{Recall (\%)} & $\mathbf{F_1}$ (\%) & \textbf{Acc (\%)}       \\ \hline

\multicolumn{1}{|c|}{\textbf{CenterTrack}} & 0.83 & 69.80    &   74.62    &  72.13   &  75.33      \\ \hline
\multicolumn{1}{|c|}{\textbf{YOLOv3}} & 0.84 & 70.23     &  73.42    & 71.79   &   76.22      \\ \hline

\end{tabular}
}
\end{table}

\subsection{Results of Different Input Sequence Lengths of LSTM} 
\label{sec:appendix4}

In \cref{tab:lstm} the results for different input sequence lengths are shown, when adding one LSTM layer with hidden size 256 before the dense layer of our YOLOv5 based $16\times16$ grids model. All sequence length achieve very similar results.

\begin{table}[h]
\caption{Comparison of different input sequence lengths when using one LSTM layer. Our model uses the $16\times16$ grids. For all metrics except $D_{KL}$, a higher value indicates the better performance. ($Th$ = 0.5)}
\label{tab:lstm}
\resizebox{.6\linewidth}{!}{% <------ Don't forget this %
\begin{tabular}{c|c|c|c|c|c|c|c|}
\cline{2-8}
& \multicolumn{5}{c|}{\textbf{Object-level}} 
& \multicolumn{2}{c|}{\textbf{Pixel-level}}  \\ \cline{2-8}
& \textit{AUC} &  \textit{Prec.  (\%)} & \textit{Recall  (\%)} & \textit{$F_1$  (\%)} & \textit{Acc  (\%)} & \textit{KL} & \textit{CC}   \\ \hline
\multicolumn{1}{|c|}{\textbf{2}}  &  0.85 & 72.40   &  72.68  & 72.54  & 78.00 &  1.16  &  0.60   \\ \hline
\multicolumn{1}{|c|}{\textbf{4}}  & 0.85  &  72.58  &  73.02  & 72.80 & 78.18 & 1.16   &  0.60   \\ \hline
\multicolumn{1}{|c|}{\textbf{6}}  &  0.85 & 72.52   &  73.04  & 72.78 & 78.16 &  1.18  &   0.60  \\ \hline
\multicolumn{1}{|c|}{\textbf{8}}  & 0.85  &  73.13  & 70.44   & 71.76  & 77.83 &  1.17  &   0.60  \\ \hline
\multicolumn{1}{|c|}{\textbf{16}}  & 0.85  &  71.84  & 73.39   & 72.61 & 77.86 &  1.18  &   0.60  \\ \hline

\end{tabular}
}
\end{table}

\section{More Qualitative Results}
\label{sec:appendix5}
\subsection{LSTM}
\begin{figure*}[h]
    \centering
        \includegraphics[height=1in]{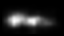}
        \includegraphics[height=1in]{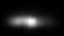}
        \includegraphics[height=1in]{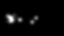}
        \includegraphics[height=1in]{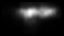}
        \includegraphics[height=1in]{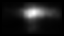}
        \includegraphics[height=1in]{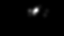}
    \caption{Comparison of predicted gaze maps without and with LSTM and ground-truth \textbf{Left}: Our prediction without LSTM; \textbf{Middle}: Our prediction with LSTM; \textbf{Right}: Ground-truth.}
    \label{fig:LSTM gaze maps}
\end{figure*}

In \cref{fig:LSTM gaze maps} there are two examples of predicted gaze maps with LSTM module (middle) in comparison with predicted gaze maps without LSTM module (left) and ground-truth (right). The LSTM module contains one layer with hidden size 256 and the input sequence length is 8. We see that the results with LSTM module enhance the prediction of the center area, which has sometimes advantages and sometimes disadvantages, thus the $AUC$ is the same (0.85). 

\subsection{BDD-A Dataset}

\begin{figure*}[h]
    \centering
        \includegraphics[width = 0.3\textwidth, height=1.1in]{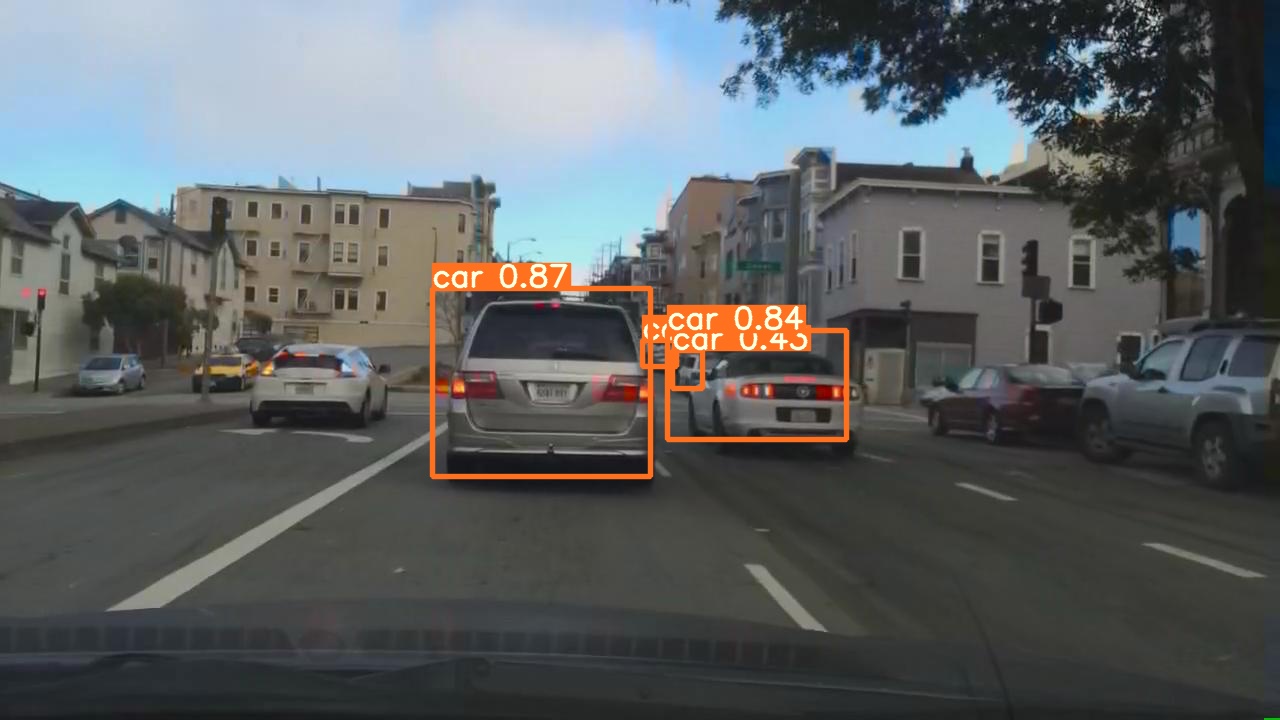}
        \includegraphics[width = 0.3\textwidth, height=1.1in]{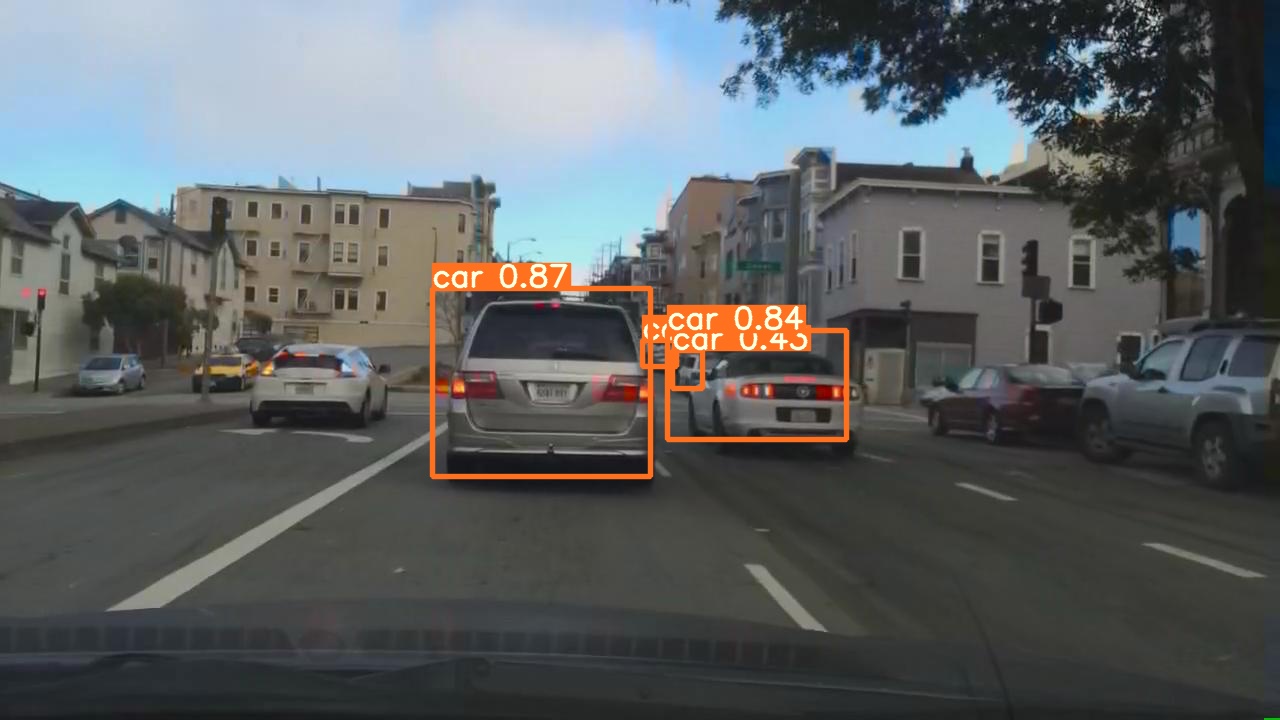}
        \includegraphics[width = 0.3\textwidth, height=1.1in]{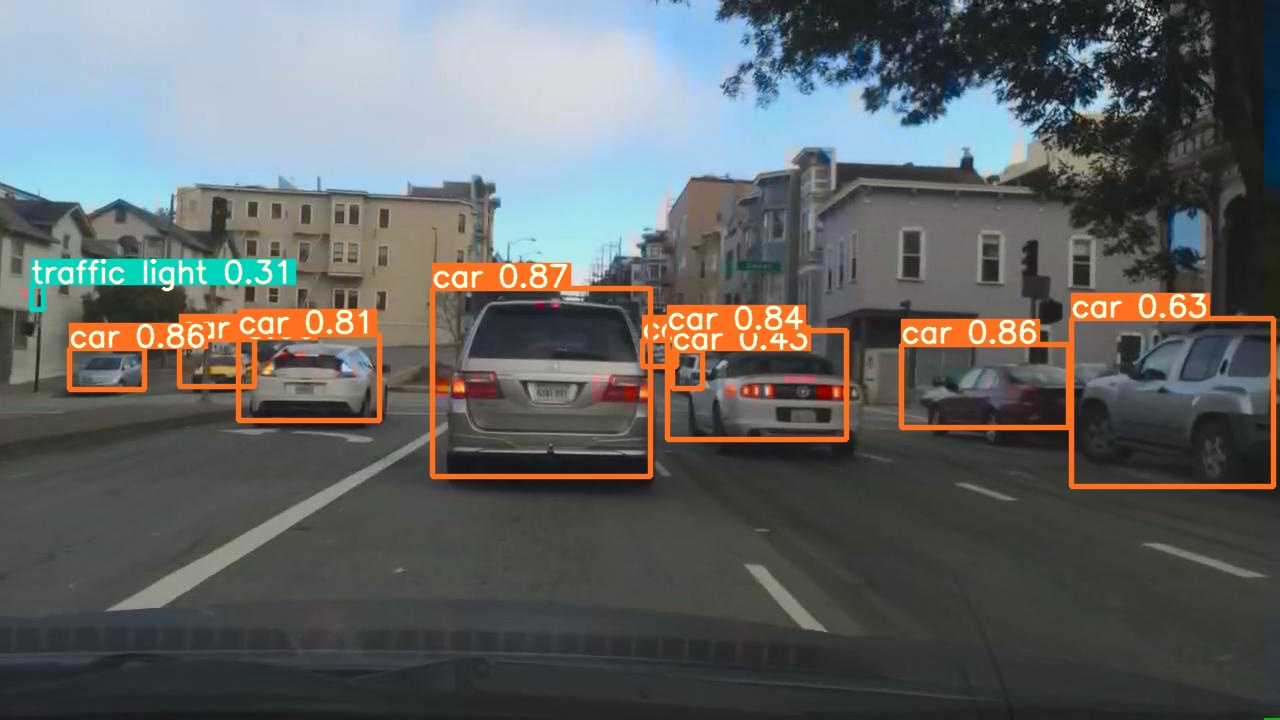}
        \includegraphics[width = 0.3\textwidth, height=1.1in]{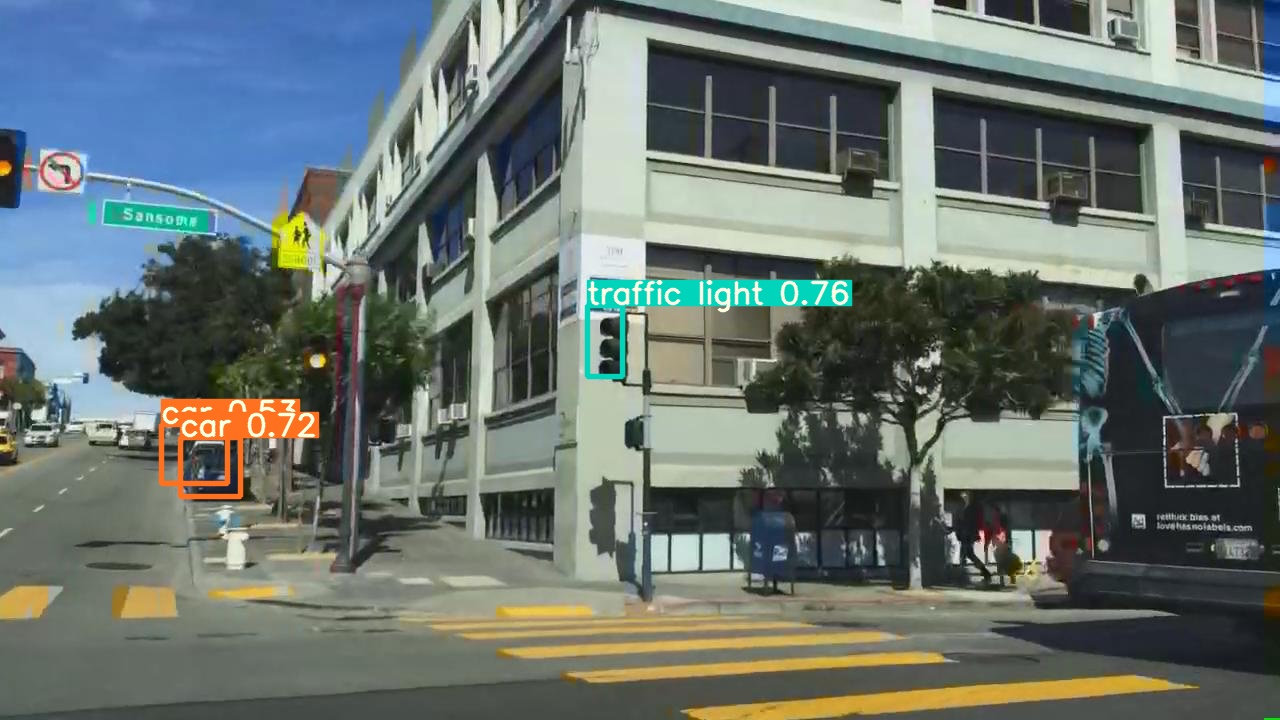}
        \includegraphics[width = 0.3\textwidth, height=1.1in]{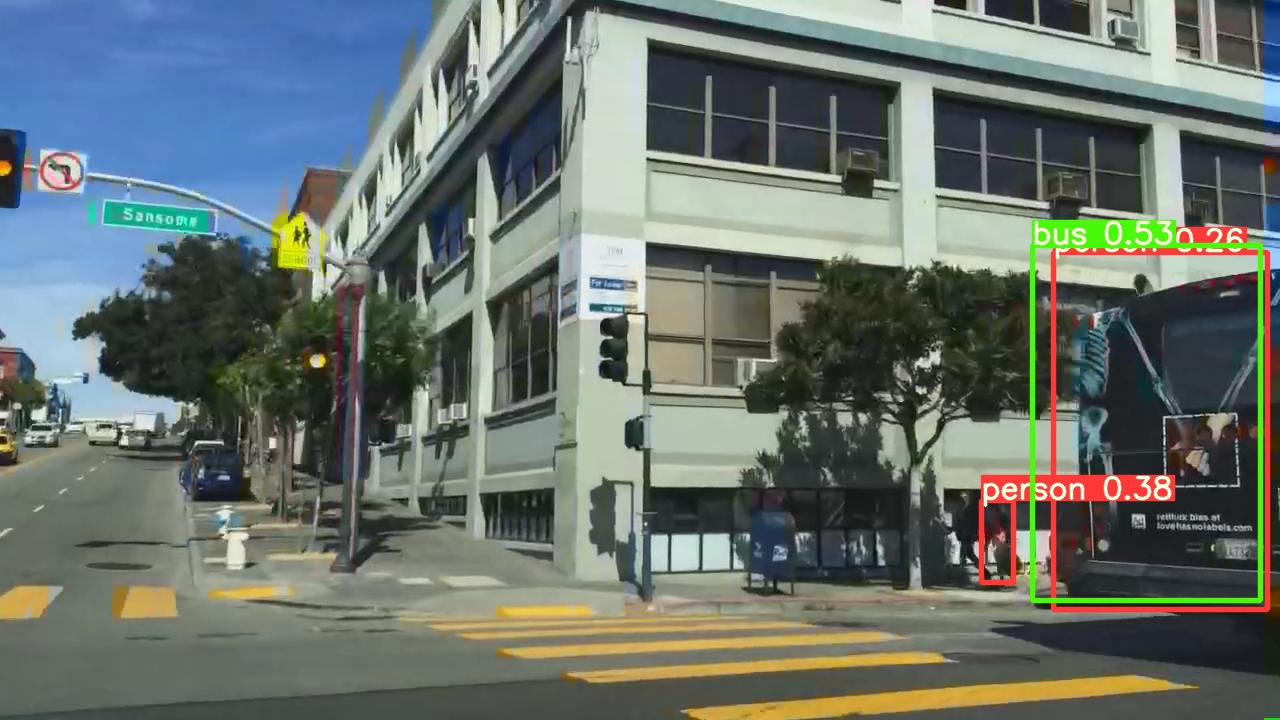}
        \includegraphics[width = 0.3\textwidth, height=1.1in]{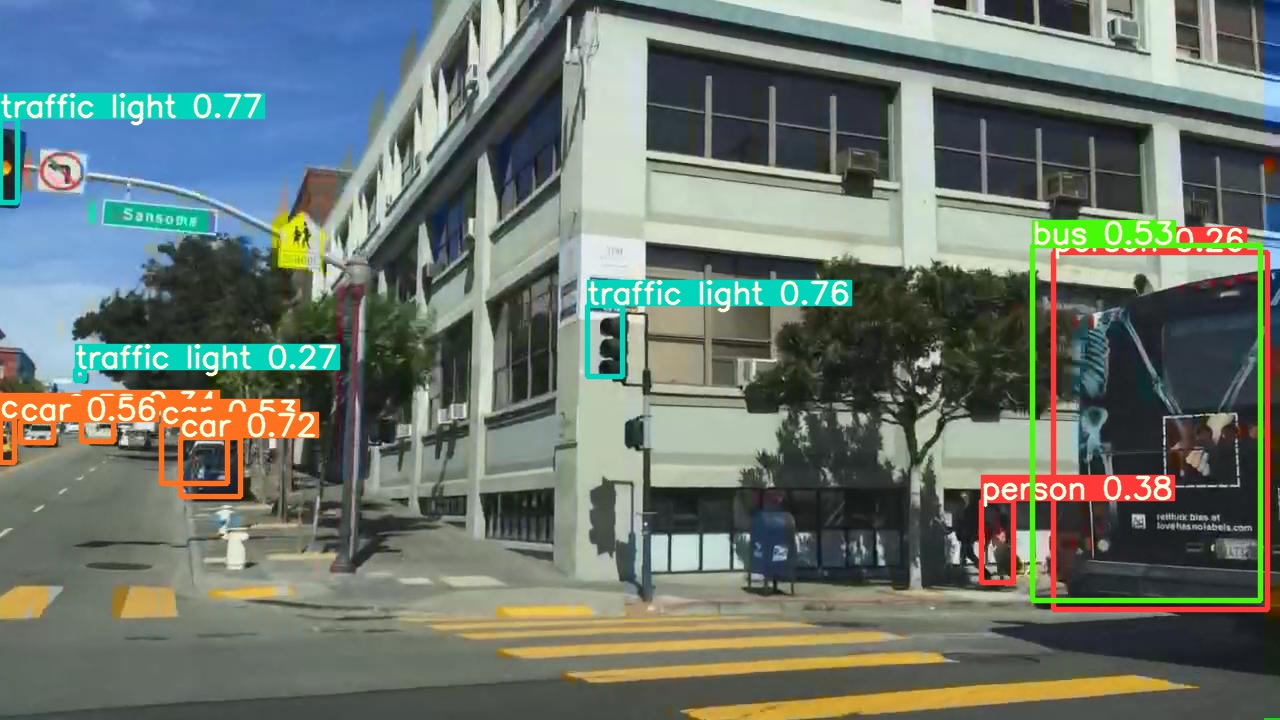}
    \caption{Comparison of our prediction, ground-truth in attention-based object detection (Th = 0.5) and not using attention-based object detection on BDD-A test set. (Second line is failed case.) \textbf{Left}: Our prediction; \textbf{Middle}: Ground-truth; \textbf{Right}: Object detection without driver attention. Better view in colors.}
    \label{fig:more bdda}
\end{figure*}

In \cref{fig:more bdda} there are two more examples of our YOLOv5 based model on BDD-A dataset. In the first row, our model predicts correctly the car on the two lanes leading straight ahead and ignoring parked cars two lanes away and another car on a turn lane. In the second row, our model predicts a traffic light in the middle of the scene, and two parked cars which could be critical if the driver would drive straight ahead. Since the driver turns left, the ground-truth covers objects on the turning road.  

\subsection{DR(eye)VE Dataset}
\begin{figure*}[h]
    \centering
        \includegraphics[width = 0.3\textwidth, height=1.1in]{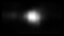}
        \includegraphics[width = 0.3\textwidth, height=1.1in]{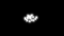}
        \includegraphics[width = 0.3\textwidth, height=1.1in]{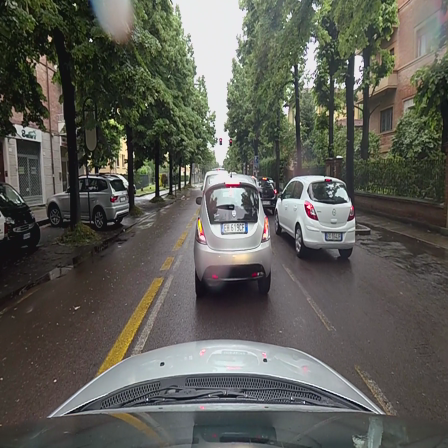}
        \includegraphics[width = 0.3\textwidth, height=1.1in]{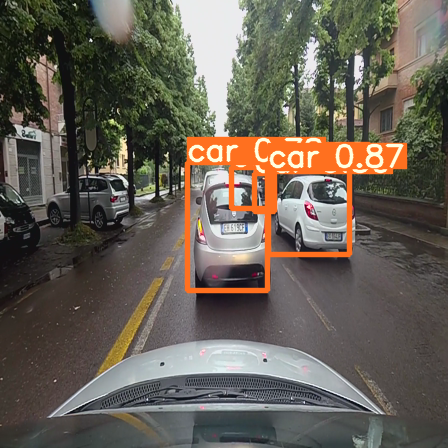}
        \includegraphics[width = 0.3\textwidth, height=1.1in]{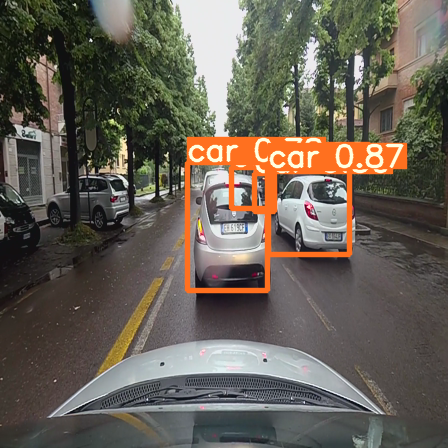}
        \includegraphics[width = 0.3\textwidth, height=1.1in]{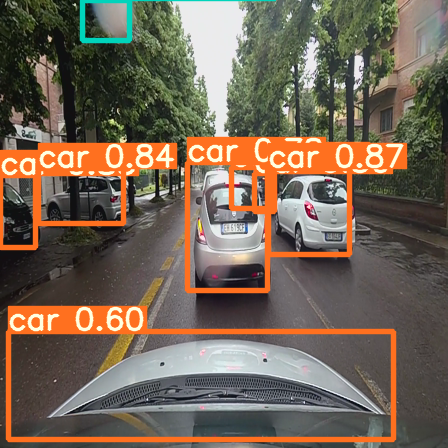}
    \caption{Comparison of our prediction, ground-truth in attention-based object detection (Th = 0.4) and not using attention-based object detection on DR(eye)VE test set.
    In the first row, from left to right: Our saliency map prediction; ground-truth saliency map; original image. In the second row, from left to right: Our prediction; ground-truth; object detection without driver attention. Better view in colors.}
    \label{fig:more dreyeve}
\end{figure*}

\begin{figure*}[h]
    \centering
        \includegraphics[width = 0.3\textwidth, height=1.1in]{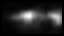}
        \includegraphics[width = 0.3\textwidth, height=1.1in]{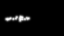}
        \includegraphics[width = 0.3\textwidth, height=1.1in]{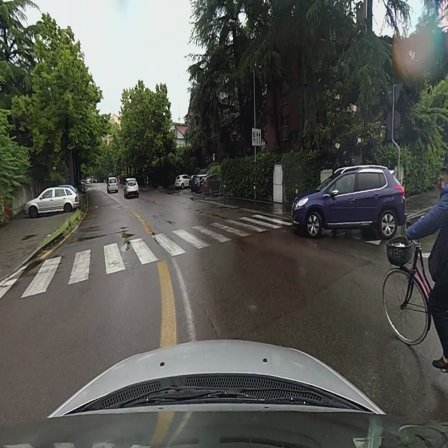}
        \includegraphics[width = 0.3\textwidth, height=1.1in]{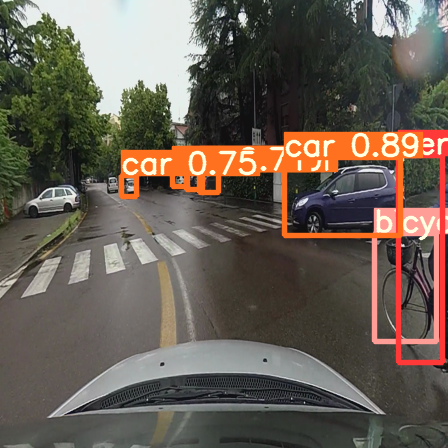}
        \includegraphics[width = 0.3\textwidth, height=1.1in]{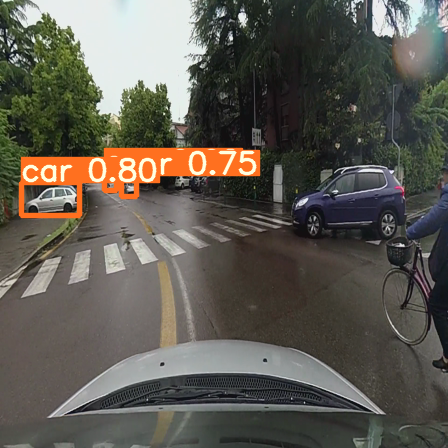}
        \includegraphics[width = 0.3\textwidth, height=1.1in]{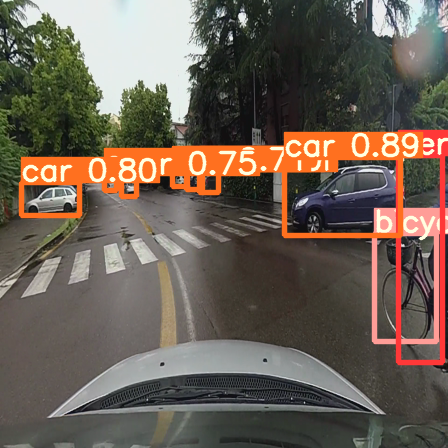}
    \caption{Comparison of our prediction, ground-truth in attention-based object detection (Th = 0.4) and not using attention-based object detection on DR(eye)VE test set. (Failed case.) In the first row, from left to right: Our saliency map prediction; ground-truth saliency map; original image. In the second row, from left to right: Our prediction; ground-truth; object detection without driver attention. Better view in colors.}
    \label{fig:more dreyeve failed 2}
\end{figure*}

\cref{fig:more dreyeve} and \cref{fig:more dreyeve failed 2} are two more examples of predicted objects with our YOLOv5 based model on DR(eye)VE dataset. In \cref{fig:more dreyeve} we see that our model predicts correctly the cars on the road and ignores the parked cars two lanes away. In \cref{fig:more dreyeve failed 2} our model predicts the cyclist next to the vehicle and a car waiting to the right, while the ground-truth focuses objects which the driver will pass later. One reason could be that the driver sees the objects next to him with peripheral view. 
%\begin{figure*}[h]
%    \centering
%        \includegraphics[width = 0.3\textwidth, height=1.1in]{images/dreyeve_qualitative/72_image-1653_gazemap_yolo5.png}
%        \includegraphics[width = 0.3\textwidth, height=1.1in]{images/dreyeve_qualitative/72_image-1653_gazemap_gt .png}
%        \includegraphics[width = 0.3\textwidth, height=1.1in]{images/dreyeve_qualitative/72_image-1653_orig.png}
%        \includegraphics[width = 0.3\textwidth, height=1.1in]{images/dreyeve_qualitative/72_image-1653_yolo5_th04.png}
 %       \includegraphics[width = 0.3\textwidth, height=1.1in]{images/dreyeve_qualitative/72_image-1653_gt.png}
 %       \includegraphics[width = 0.3\textwidth, height=1.1in]{images/dreyeve_qualitative/72_image-1653.png}
%    \caption{Dreyeve bad TH=0.4, gt focus mainly person, pred focus also traffic light}
 %   \label{fig:more dreyeve failed}
%\end{figure*}

\end{document}